\documentclass{article} 
\usepackage{iclr2026_conference,times}

\usepackage{amsmath,amsfonts,bm}

\def\eqref#1{equation~\ref{#1}}

\def\1{\bm{1}}

\DeclareMathAlphabet{\mathsfit}{\encodingdefault}{\sfdefault}{m}{sl}
\SetMathAlphabet{\mathsfit}{bold}{\encodingdefault}{\sfdefault}{bx}{n}

\usepackage{hyperref}
\usepackage{url}
\usepackage{graphicx}
\usepackage{amsmath}
\usepackage{amsthm}
\usepackage{amssymb}
\usepackage{booktabs}
\usepackage{multirow,multicol}
\usepackage{algorithm}
\usepackage{algorithmic}
\usepackage{placeins}
\usepackage{xcolor}
\usepackage{enumitem}
\setlist[itemize]{leftmargin=3.5mm, itemsep=0.1em, topsep=0.1em}
\setlist{nolistsep}
\usepackage{tabularx}

\usepackage{caption}
\DeclareCaptionFont{ninept}{\fontsize{9pt}{11pt}\selectfont}
\title{Evolving Safety Landscape of Multi-modal Large Language Models: A Survey of Emerging Threats and Safeguards}

\author{Xi Li\textsuperscript{1}\thanks{Corresponding author: \texttt{xli7@uab.edu}} \quad
Shu Zhao\textsuperscript{2} \quad
Xiaohan Zou\textsuperscript{3} \quad
Fei Zhao\textsuperscript{1} \quad
Fuxiao Liu\textsuperscript{2} \\
\textbf{Yusen Zhang}\textsuperscript{4} \quad
\textbf{Cheng Han}\textsuperscript{5} \quad
\textbf{Yushun Dong}\textsuperscript{6} \quad
\textbf{Jiaqi Wang}\textsuperscript{7} \\
\textsuperscript{1}University of Alabama at Birmingham, 
\textsuperscript{2}NVIDIA,
\textsuperscript{3}Penn State University, 
\textsuperscript{4}Columbia University, \\
\textsuperscript{5}University of Missouri-Kansas City,
\textsuperscript{6}Florida State University,
\textsuperscript{7}Auburn University
}

\iclrfinalcopy 
\begin{document}

\maketitle

\begin{abstract}

Multi-modal large language models (MLLMs) integrate heterogeneous modalities through modality alignment and fusion, enabling stronger understanding and reasoning. However, this architectural shift reshapes the safety landscape of machine learning. Increased model complexity and cross-modal interactions give rise to novel threats, including compromised modality integration, modality misalignment, and fused safety risks, reflecting shifts in threat modeling beyond uni-modal assumptions. These shifts, in turn, impose new constraints on safety solutions not captured by existing frameworks rooted in uni-modal learning.
Motivated by these challenges, this survey provides a systematic analysis of the evolving safety landscape of MLLMs.
We first propose a multimodal grounded taxonomy of safety threats and analyze shifts in threat models, covering adversarial attacks, data poisoning, jailbreaks, and hallucinations. 
We then summarize updated safety assumptions and organize recent advances in MLLM safety strategies accordingly. 
Finally, we discuss open challenges and future directions to inform the development of more principled and scalable safety mechanisms for multimodal systems.

\end{abstract}

\vspace{-0.1in}
\section{Introduction}\label{sec:intro}
\vspace{-0.1in}

Modern multi-modal learning employs large models, such as large language models (LLMs), to integrate diverse modalities (e.g., text, image, audio, and video) for enhanced understanding and decision-making~\citep{BLIP,Flamingo,LLAVA,GPT,DeepSeek,Llama3}, enabling applications in healthcare, autonomous driving, and legal decisions. Its foundation lies in modality alignment, which maps heterogeneous features into a shared representation space, and modality fusion, which integrates aligned information for more comprehensive and accurate  reasoning~\citep{yin2024survey,wang2023large,zhu2023multimodal,xu2023multimodal,baltruvsaitis2018multimodal}.

With the shift toward multi-modal architectures, the safety landscape of machine learning is undergoing a significant transformation. The unique properties of multi-modal learning introduce new safety challenges: (1) additional modalities expand the attack surface by introducing their own vulnerabilities; (2) cross-modal interactions may exhibit semantic inconsistencies and can be exploited to induce adversarial misalignment; and (3) the fusion process can be exploited, where malicious signals that appear benign in isolation may trigger harmful behavior when combined. These emerging threats reflect a shift in threat modeling, where attackers require only partial system access and can exploit surfaces beyond the input space, leading to cross-modal vulnerabilities.

These shifts further introduce new constraints on safety solutions that go beyond classic uni-modal assumptions. For example, defenses can no longer presume which modality is compromised or the specific type of threat, and must instead align with core multi-modal objectives such as modality alignment and fusion, while also monitoring internal model stages. Addressing these constraints calls for safety mechanisms that are both effective and cost-efficient, including safety-aware fine-tuning that preserves inter-modal coherence, safety-integrated preference optimization, and training-free inference-time solutions.

These changes motivate a comprehensive survey on the safety of multi-modal large language models (MLLMs). Recent surveys, summarized in Table~\ref{tab:prior_surveys}, provide valuable overviews of attacks, defenses, and evaluation protocols \citep{IJCAI_survey,llms-mllms,10831129,FCS_VLM_security,jin2024jailbreakzoo,ye2025surveysafetyVLM,TNNLS_survey}, but largely adopt uni-modal taxonomies that organize studies by training/test threats or by separating vision-side and text-side attacks. 
Such perspectives may not fully capture risks unique to multi-modal systems, particularly those arising from cross-modal interactions. 
Moreover, prior surveys largely enumerate threats and defenses, with limited discussion of paradigm shifts in multi-modal safety, such as evolving threat models and safety constraints, offering limited guidance for future research on multi-modal safety.

\begin{figure*}[t]
\vspace{-0.2in}
    \centering
    \includegraphics[width=.98\linewidth]{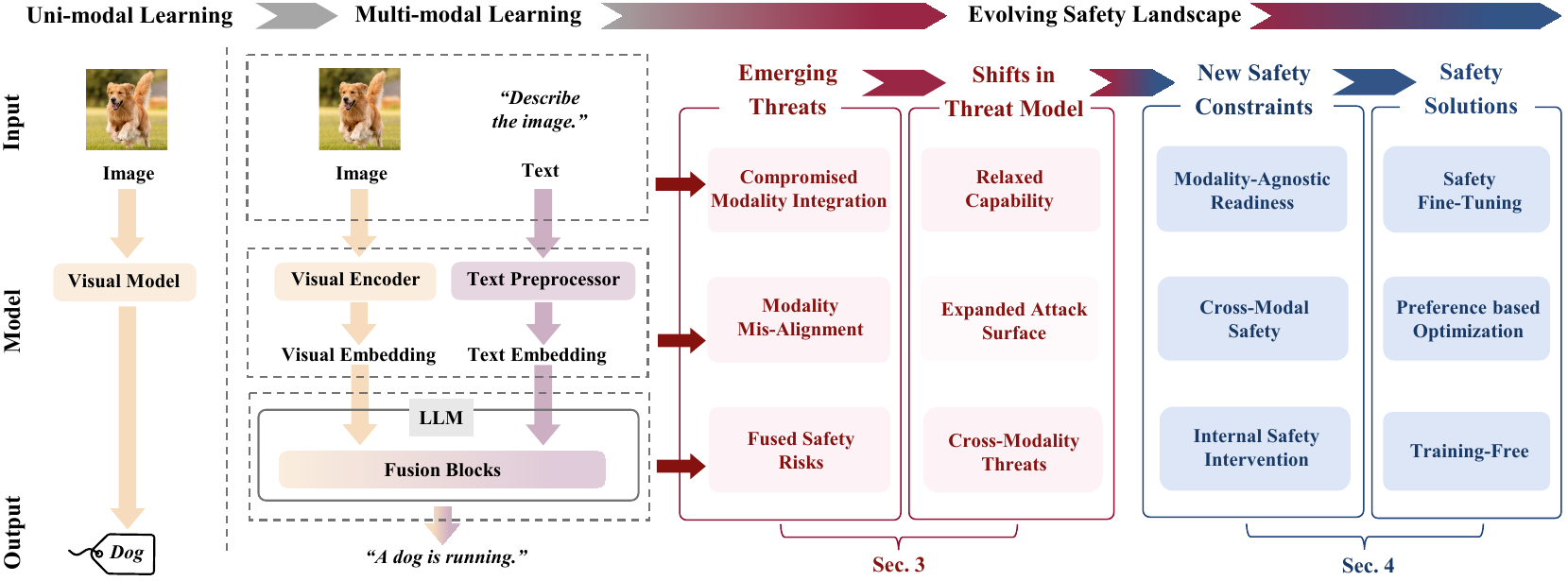}
    \vspace{-0.05in}
    \caption{A multi-modal perspective on the evolving safety landscape, illustrated with Vision-Language LLM.}
    \label{fig:Overview}
\vspace{-0.2in}
\end{figure*}

Motivated by this gap, this survey provides a systematic analysis of the evolving safety landscape shaped by MLLMs (Figure~\ref{fig:Overview}), with the core contributions outlined as follows:

\begin{itemize}
    \item We propose a new taxonomy of emerging safety threats in MLLMs arising from complex model architectures and cross-modal interactions, including compromised modality integration, cross-modal misalignment, and risks at the fusion stage. We further summarize the evolving threat modeling reflected by these threats.

    \item We summarize updated assumptions required for safety solutions in multi-modal settings and categorize recent advances in multi-modal safety within this framework, including supervised safety fine-tuning, preference-based optimization, and training-free methods.

    \item As safety research shifts from uni-modal to multi-modal systems, we highlight the need to address the increased structural complexity and interaction dynamics of MLLMs, and outline key directions for future research.
    
\end{itemize}


Although we discuss MLLM safety broadly, much of the literature focuses on vision-language models (VLMs), the most widely studied instantiation. Nevertheless, our key concepts generalize to MLLMs with additional modalities (e.g., audio, video, and sensory inputs).
The remainder of the paper is organized as follows: Section~\ref{sec:bg} reviews uni-modal safety foundations; Section~\ref{sec:threats} presents MLLM characteristics, a taxonomy of emerging threats, and evolving threat models; Section~\ref{sec:solutions} summarizes updated safety assumptions and recent safety solution advances; Section~\ref{sec:future} discusses future directions; and Section~\ref{sec:conclusion} concludes.

\FloatBarrier

\begin{table*}[t]
    \centering
    \vspace{-0.2in}
    \resizebox{0.96\linewidth}{!}{
    \begin{tabular}{c|c|c|c|c|c|c|c|c|c}
    \toprule
    \multirow{2}{*}{\textbf{Survey}} & \multirow{2}{*}{\textbf{Venue}} & \multicolumn{4}{c|}{\textbf{Threats}} & \multirow{2}{*}{\textbf{Solutions}} & \multirow{2}{*}{\textbf{\shortstack{MM \\ Tax.}}} & \multirow{2}{*}{\textbf{\shortstack{Safety \\ Evo}}} & \multirow{2}{*}{\textbf{Key Contributions}}  \\
    \cline{3-6}
    & & \textbf{A} & \textbf{P} & \textbf{J} & \textbf{H} & & & \\
    \hline
    \citet{IJCAI_survey} & IJCAI & \checkmark & & \checkmark &  & \checkmark &  &  &  Concise overview and basic categorization of VLM security \\
    \hline
    \citet{10831129} & SMC & \checkmark & \checkmark & & & \checkmark & & & Overview of image-induced attacks and defenses \\
    \hline
    \citet{FCS_VLM_security} & FCS & & \checkmark & \checkmark & \checkmark & \checkmark & &  & VLM security: attacks, defenses, and evaluation\\
    \hline
    \citet{jin2024jailbreakzoo} & arXiv & & &  \checkmark & & \checkmark & & & Comprehensive survey of jailbreak in LLMs and MLLMs \\
    \hline
    \citet{llms-mllms} & EMNLP & & &  \checkmark & & \checkmark & & & Eovolving Jailbreak landscape from LLMs to MLLMs \\
    \hline
    \multirow{2}{*}{\citet{ye2025surveysafetyVLM}} & \multirow{2}{*}{arXiv} & \multirow{2}{*}{\checkmark} & \multirow{2}{*}{\checkmark} & \multirow{2}{*}{\checkmark} & & \multirow{2}{*}{\checkmark} & \multirow{2}{*}{} & & Lifecycle-based classification of VLM security \\
    & & & & & & & & & across attacks, defenses, and evaluation \\
    \hline
    \citet{TNNLS_survey} & TNNLS & \checkmark & \checkmark & \checkmark & &  &  &  & Overview of VLM attacks and evaluation datasets/models \\
    \hline
    \multirow{2}{*}{\textbf{Ours}} & \multirow{2}{*}{-} & \multirow{2}{*}{\checkmark} & \multirow{2}{*}{\checkmark} & \multirow{2}{*}{\checkmark} & \multirow{2}{*}{\checkmark} & \multirow{2}{*}{\checkmark} & \multirow{2}{*}{\checkmark} & \multirow{2}{*}{\checkmark} & MM-grounded taxonomy of MLLM safety and \\
    & & & & & & & & & in-depth analysis of safety landscape shifts \\
    \bottomrule
    \end{tabular}
    }
    \vspace{-0.05in}
    \caption{Recent surveys on MLLM safety. A: adversarial attacks; P: poisoning attacks; J: jailbreak attacks; H: hallucination; MM Tax.: MM-grounded taxonomy; Safety Evo: safety landscape evolution; VLM: vision-language model.}
    \label{tab:prior_surveys}
    \vspace{-0.2in}
\end{table*}

\vspace{-0.1in}
\section{Existing Safety Landscape}\label{sec:bg}
\vspace{-0.1in}

To provide necessary context, we review the well-established safety landscape shaped primarily by uni-modal learning.
We follow standard attacker access definitions: white-box (full model knowledge), black-box (query-only access), and gray-box (partial knowledge, e.g., training data, architecture without parameters, or logits).

\vspace{-0.1in}
\subsection{Adversarial Attacks}
\vspace{-0.1in}

Adversarial attacks seek to induce incorrect outputs at inference time by adding imperceptible perturbations to inputs. Attackers manipulate test-time inputs without access to training data, operating in either black-box \citep{ZOO} or white-box \citep{FGSM,PGD,CW} settings. Such perturbations are typically generated via optimization-based methods that either target specific incorrect outputs or maximize prediction error, while satisfying norm-based constraints (e.g., $l_\infty$, $l_2$) to preserve perceptual imperceptibility, especially for images.


Defenses aim to enhance robustness against subtle perturbations, typically assuming full access to the victim model, sometimes with a small clean validation set. Adversarial training \cite{adv_training} improves robustness by iteratively generating adversarial examples and training the model to correctly classify them, while random smoothing \cite{random_smoothing,LiCWC19} injects Gaussian noise and averages predictions to smooth decision boundaries against small perturbations.

\vspace{-0.1in}
\subsection{Data Poisoning}
\vspace{-0.1in}

Data poisoning manipulates model behavior by injecting malicious training samples, assuming attacker access to the training set. 
Attacks range from label flipping \cite{XiaoXE12,ZhangCZL21} that degrades overall performance to backdoor attacks that induce targeted misbehavior via trigger patterns while preserving clean accuracy. 
Triggers can take various forms, including subtle perturbations \cite{WaNet,input-aware,HiddenTrigger}, imperceptible visual patterns in images \cite{BadNet,Targeted-Backdoor}, and specific tokens \cite{BadWord}, sentences \cite{AddSent}, or writing styles \cite{QiCZLLS21} in text.

Poisoning defenses aim to reduce the impact of poisoned data while preserving model utility. Pre-training defenses detect and remove suspicious samples from the training set \cite{SS,PaudiceML18}, training-time defenses leverage white-box access to select reliable samples or reduce poisoning effects during optimization \cite{DiakonikolasKK019,trim_loss,li2021anti,DPA,FA}, and post-training defenses use white-box access and a small clean set to detect poisoned models \cite{NC,UNICORN}, identify poisoned inputs \cite{STRIP,in-flight}, or enforce benign behavior \cite{FP,NAD,hypergrad,BNA}.

\vspace{-0.1in}
\subsection{Jailbreak}
\vspace{-0.1in}

Jailbreak attacks bypass LLM safety by crafting prompts that elicit unauthorized rather than misleading outputs. White-box attacks optimize adversarial prefixes or suffixes \cite{abs-2307-15043,JonesDRS23,abs-2310-15140}, 
Gray-box attacks exploit logit access to manipulate prompts and token selection \cite{GuoYZQ024,abs-2401-17256}, or perform limited retraining via small-scale poisoning \cite{Qi0XC0M024,ZhanFBGHK24,abs-2310-02949}.
and black-box attacks leverage role-playing or contextual reasoning to bypass safety alignment \cite{abs-2305-14950,abs-2310-06387,abs-2311-03191}. Other methods include LLM-generated adversarial prompts \cite{Deng_2024,liu2024autodan} and niche-language evasion \cite{YuanJW0H0T24}.


Jailbreak defenses seek to maintain safety alignment. Black-box defenses filter or preprocess adversarial prompts to neutralize harmful intent \cite{abs-2308-14132,abs-2309-00614}, while white-box defenses impose stronger guardrails via safety fine-tuning \cite{0001SARJH024,DengWFDW023}, reinforcement learning from human feedback (RLHF) \cite{Ouyang0JAWMZASR22}, or model self-correction to reduce unsafe outputs \cite{SunSZZCCYG23}.

\vspace{-0.1in}
\subsection{Hallucination}
\vspace{-0.1in}

Hallucination refers to the generation of confident but factually incorrect or unsupported information \cite{huang2025survey}, arising from limitations in training data, objectives, or decoding. It is mainly a reliability issue rather than a security issue involving active attackers. Common causes include noisy or biased data \cite{bender2021dangers,sheng2020towards,wallace2019universal}, overfitting to spurious language patterns \cite{mckenna2023sources,wang2021identifying,tu2020empirical}, and limited uncertainty modeling \cite{farquhar2024detecting}.


To mitigate hallucination, some works improve data quality through filtering, deduplication, and high-quality structured or instructional data \citep{penedo2023refinedweb,lee2021deduplicating,guu2020retrieval,raffel2020exploring}. Other works adopt RLHF, contrastive learning, factual-consistency losses, or chain-of-thought supervision \citep{ziegler2019fine,sun2023contrastive,wei2022chain}. Additional works employ retrieval-augmented generation (RAG), prompt constraints, and post-hoc verification using confidence estimation or external tools \citep{lewis2020retrieval,izacard2020leveraging,zhou2022least,beurer2023prompting,gao2023pal,farquhar2024detecting,li2023making}.


\vspace{-0.1in}
\section{Emerging Multi-Modal Safety Threats}
\label{sec:threats}
\vspace{-0.1in}

\begin{figure*}[t]
    \centering
    \vspace{-0.2in}
    \includegraphics[width=0.98\linewidth]{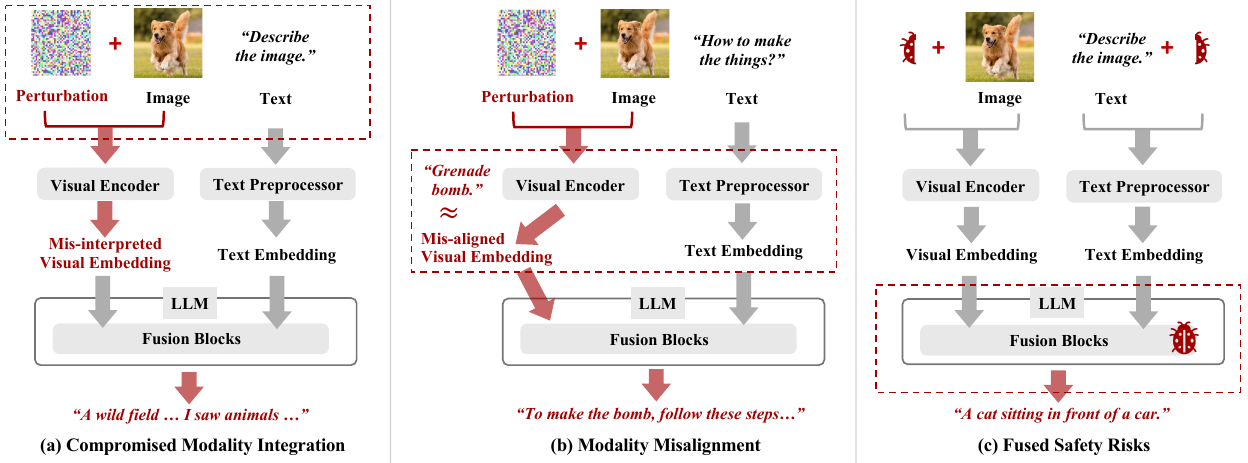}
    \vspace{-0.05in}
    \caption{Illustration of (a) compromised visual integration \citep{GaoBGX00024}; (b) targeted misalignment for jailbreaks \citep{jailbreak_adv}; (c) fused backdoor attacks \citep{WalmerSSSJ22}.}
    \label{fig:threats}
    \vspace{-0.2in}
\end{figure*}

To integrate diverse data sources and enable cross-modal reasoning, MLLMs rely on two core mechanisms: modality alignment and modality fusion (Fig.~\ref{fig:Overview}). 
Modality alignment projects heterogeneous inputs into a shared embedding space using modality-specific encoders while preserving essential features, whereas modality fusion combines aligned representations (e.g., via cross-attention) to capture cross-modal semantics and support downstream reasoning tasks. 
However, the inherent complexity of these mechanisms and dynamic cross-modal interactions give rise to a new set of safety threats differ fundamentally from uni-modal settings. 
To characterize these emerging threats, we propose a new taxonomy of multi-modal safety risks (Fig.~\ref{fig:threats}): Compromised Modality Integration, Modality Misalignment, and Fused Safety Risks.


\vspace{-0.1in}
\subsection{Compromised Modality Integration}
\vspace{-0.1in}
\label{sec:threats_CMI}

When multiple modalities are integrated, threats may originate from \textbf{individual modalities}, where manipulation of one or more inputs propagates through the integration process and compromises overall model behavior.
We refer to such threats as Compromised Modality Integration (Fig.~\ref{fig:threats} (a)).

Adversarial attacks in MLLMs primarily target the vision encoder by inserting visual perturbations that disrupt its representations. These errors then propagate through the alignment and fusion processes, compromising the behavior of the entire system.
For example, 
\citet{LuoGL024} and \citet{Schlarmann023} craft visual perturbations that induce MLLMs to generate attacker-specified text.
Several works
\citet{0003YZDZLCWM23,WangDZQLFWL24,WangLQCJX24} maximize the embedding distance between clean and perturbed images, distorting the model’s understanding.
\citet{GaoBGX00024} introduce perturbations that delay the end-of-sequence token, increasing uncertainty in output length.
\citet{BaileyO0E24} show that adversarial images can bypass safety filters, leak context, and trigger misleading.
\citet{fu2023misusing} demonstrate that subtle perturbations can covertly induce MLLM agents to perform malicious actions, such as leaking conversations.
\citet{cai2025imperceptible} boost transferability by applying visual transformations to ensure all perturbed variants fool the model.
\citet{zhang2025anyattack} improve scalability, by using a self-supervised setup where a random image generates perturbations guided by the original.
\citet{zhang2024b} propose a benchmark of perturbed inputs that consistently trigger low-quality MLLM responses.

\citet{LiuZ25} poison the visual encoder to map backdoored images to a target representation, inducing MLLM outputs aligned with the target content.
MLLMs inherit jailbreak vulnerabilities from their LLM core, making them susceptible to harmful or unauthorized outputs \cite{liu2024autodan,YuLLCXZ24,Deng_2024}.
Attackers can manipulate non-text modalities, such as image, to bypass safety mechanisms of LLM, triggering jailbreak responses with benign text and malicious visuals.
For example, \citet{QiHP0WM24} optimize a visual adversarial example to universally jailbreak an aligned LLM.
FigStep~\cite{FigStep} embeds rephrased jailbreak prompts within images to deceive the model. 
\citet{jailbreak_adv} craft visual perturbations that bring visual signals close to jailbreak content in the embedding space.

During multi-modal integration, perceptual flaws in individual modalities can propagate and lead to system-wide hallucinations. For instance, low-resolution images may obscure small but critical objects, or speech recognition may omit key words. 
LLaVA-Next \cite{liu2024llavanext} shows that higher-resolution inputs reduce hallucinations, while Prismatic \cite{karamcheti2024prismatic} leverages expert encoders to preserve fine-grained details. When such information is lost, the model tends to rely on pretraining priors to ``fill in the gaps,'' often resulting in hallucinated content.

\begin{table*}[t]
    \centering
    \vspace{-0.2in}
    \resizebox{0.98\linewidth}{!}{
    \begin{tabular}{c|c|c|c|c|c|c}
    \toprule
        \multirow{2}{*}{\textbf{Category}} & \multirow{2}{*}{\textbf{Sub-Category}} & \multirow{2}{*}{\textbf{Paper}} & \multicolumn{3}{c|}{\textbf{Threat Model}} & \multirow{2}{*}{\textbf{Year \& Venue}} \\
        \cline{4-6}
        & & & \textbf{Cap. 1} & \textbf{Cap. 2} & \textbf{Cap. 3} & \\
        \hline
        \multirow{10}{*}{\textbf{\shortstack{Compromised \\ Modality \\ Integration}}} 
            & \multirow{4}{*}{Adversarial Attacks} & CroPA\cite{LuoGL024} & \checkmark & & & ICLR'24 \\
            &                                      & Verbose Images\cite{GaoBGX00024} & \checkmark & & & ICLR'24 \\
            &                                      & Image Hijacks\cite{BaileyO0E24} & \checkmark & & \checkmark & ICML'24 \\
            &                                      & AnyAttack\cite{zhang2025anyattack} & \checkmark & & & CVPR'25 \\
        \cline{2-7}
            & Poisoning & BadVision\cite{LiuZ25} & \checkmark & & & CVPR'25 \\
        \cline{2-7}
            & \multirow{3}{*}{Jailbreak} & Adversarial Vision Jailbreaks\cite{QiHP0WM24} & \checkmark & & \checkmark & AAAI'24 \\
            &                            & Jailbreak In Pieces\cite{jailbreak_adv} & \checkmark & \checkmark & \checkmark & ICLR'24 \\
            &                            & FigStep\cite{FigStep} & \checkmark & & \checkmark & AAAI'25 \\
        \cline{2-7}
            & \multirow{2}{*}{Hallucination} & LLaVA-Next\cite{liu2024llavanext} & - & - & - & arXiv’24 \\
            &                                & Prismatic VLMs\cite{karamcheti2024prismatic} & - & - & - & ICML’24 \\
        \hline
        \multirow{16}{*}{\textbf{\shortstack{Modality \\ Mis-alignment}}} 
        & \multirow{3}{*}{\shortstack{Untargeted \\ Mis-alignment}} & Pandora’s Box\cite{LiuYQ0FT0024} & & \checkmark & & NeurIPS’24 \\
        &                                           & Break the Visual Perception\cite{WangLQCJX24} & \checkmark & \checkmark & & ACM MM’24 \\
        &                                           & TMM Attack\cite{WangDZQLFWL24} & & \checkmark & & IEEE S \& P'24 \\
        \cline{2-7}
        & \multirow{9}{*}{\shortstack{Targeted \\ Mis-alignment}} & VLM Adversarial Robustness\cite{ZhaoPDYLCL23} & & \checkmark & & NeurIPS’23 \\
        &                                         & Adversarial Illusions\cite{bagdasaryan2024adversarial} & & \checkmark & & USENIX Security’24 \\
        &                                         & Shadowcast\cite{XuYSSW0GH24} & & \checkmark & & NeurIPS’24 \\
        &                                         & BadCLIP\cite{BaiGMX0024} & & \checkmark & & CVPR’24 \\
        &                                         & VLM Safety Evaluation\cite{tu2024many} & & \checkmark & & ECCV’24 \\
        &                                         & TrojVLM\cite{TrojVLM} & & \checkmark & & ECCV’24 \\
        &                                         & Revisit VLM Backdoor\cite{LiangLPDLZCT25} & & \checkmark & & CVPR’25 \\
        &                                         & VLOOD\cite{LyuYG00YHL025} & & \checkmark & & ICLR’25 \\
        &                                         & BadSem\cite{BadSem} & & \checkmark & & arXiv’25 \\
        \cline{2-7}
        & \multirow{4}{*}{\shortstack{Semantic \\ Inconsistency}} & Minigpt-4\cite{zhu2023minigpt} & - & - & - & arXiv’23 \\
        &                                         & Blip-2\cite{li2023blip} & - & - & - & ICML’23 \\
        &                                         & InstructBLIP\cite{dai2023instructblipgeneralpurposevisionlanguagemodels} & - & - & - & arXiv’23 \\
        &                                         & mPLUG-Owl2\cite{ye2024mplug} & - & - & - & CVPR’24 \\
        \hline
        \multirow{4}{*}{\textbf{\shortstack{Fused \\ Safety Risks}}} 
        & \multirow{3}{*}{\shortstack{Fusion-triggered \\ threats}} & Dual-Key Backdoor\cite{WalmerSSSJ22} & & \checkmark & & CVPR’22 \\
        &                                           & AnyDoor Attack\cite{abs-2402-08577} & & \checkmark & \checkmark & arXiv’24 \\
        &                                           & ImgTrojan\cite{ImgTrojan} & \checkmark & \checkmark & \checkmark & NAACL’25 \\
        \cline{2-7}
        & Modality Conflicts & Curse of Multi-Modalities\cite{leng2024} & - & - & - & arXiv’24 \\
    \bottomrule
    \end{tabular}
    }
    \vspace{-0.05in}
    \caption{Representative studies of multi-modal safety threats under the new taxonomy. Cap. 1: relaxed capability; Cap. 2: expanded attack surface; Cap. 3: cross-modality threats.}
    \label{tab:sum}
    \vspace{-0.2in}
\end{table*}

\vspace{-0.1in}
\subsection{Modality Misalignment}
\vspace{-0.1in}
\label{sec:threats_MM}

Adversaries can manipulate \textbf{cross-modal embeddings} to disrupt semantic or structural alignment between modalities, misleading the model during inference (Fig.~\ref{fig:threats}(b)).
Adversarial misalignment can be:
(1) \textit{Untargeted}, where the perturbed modality’s embedding diverges from clean modalities; or
(2) \textit{Targeted}, where embeddings are manipulated to mimic harmful representations.
Misalignment may also arise naturally due to inherent inconsistencies across modalities.

For \textit{untargeted} misalignment, existing works generally maximize cross-modal discrepancy/distance in the shared embedding space, thereby disrupting alignment between modalities. 
For example, \citet{LiuYQ0FT0024} generate a universal adversarial patch by minimizing cosine similarity between visual and textual embeddings.
\citet{WangLQCJX24} introduce visual perturbations that disrupt relations among visual tokens or weaken global visual–text semantic alignment. 
\citet{WangDZQLFWL24} focus on suppressing features that promote modality consistency while amplifying those that increase cross-modal discrepancy.

For \textit{targeted} misalignment, \citet{ZhaoPDYLCL23} align adversarial image embeddings with (1) the target text, (2) embeddings of corresponding target images, or (3) model outputs associated with the target text.
\citet{jailbreak_adv} optimize visual perturbations to mimic embeddings of harmful content (e.g., OCR-decoded jailbreak prompts), enabling cross-modal jailbreaks.
Similarly, \citet{QiHP0WM24} craft adversarial visuals to maximize the likelihood of harmful text generation, breaking safety alignment. 
\citet{bagdasaryan2024adversarial} and \citet{tu2024many} craft perturbations that minimize cosine similarity between the input and an unrelated adversarial target text.
Furthermore, several works \cite{lu2023set,han2023ot,huangx} improve the transferability of embedding misalignment.

Some methods induce targeted misalignment via data poisoning.
\citet{TrojVLM,LiangLPDLZCT25,LyuYG00YHL025} implant backdoors in MLLMs that trigger targeted harmful outputs when images contain backdoor patterns.
Further, \citet{XuYSSW0GH24} poison the model to align embeddings of original and destination concept images, causing it to generate text about the destination when shown the original.
\citet{BaiGMX0024} introduce sample-specific triggers and trigger-aware context prompts to shift visual embeddings toward a target class.
\citet{BadSem} exploit semantic inconsistencies between paired modalities, e.g., mismatches in color or object descriptions, to trigger backdoor behaviors.

Inherent semantic inconsistencies across modalities can naturally cause misalignment and lead to hallucinations.
Besides, modality alignment projects heterogeneous inputs into a shared embedding space, which is often anchored to text. Improper mappings may distort semantics and induce hallucinated outputs \citep{zhu2023minigpt}. 
For example, visual features from a medical scan may align with incorrect clinical concepts, causing the model to generate convincing but unfaithful outputs.
Methods like BLIP-2 \citep{li2023blip} and InstructBLIP \citep{dai2023instructblipgeneralpurposevisionlanguagemodels} use query-based feature extraction to improve cross-modal faithfulness, but fail when queries miss critical visual details.
Similarly, mPLUG-Owl2 \citep{ye2024mplug} employs modality-adaptive attention, yet misalignment persists under domain shifts or out-of-distribution inputs.

\vspace{-0.1in}
\subsection{Fused Safety Risks}
\vspace{-0.1in}
\label{sec:threats_FSR}

Fused safety risks exploit the fusion mechanism, where adversarial signals are benign in isolation but \textbf{harmful when combined} during fusion, as shown in Fig.~\ref{fig:threats} (c). These threats manifest deeper in the model pipeline and are harder to detect. 
Beyond adversarial manipulation, naturally conflicting or imbalanced cross-modal evidence can also lead to unsafe behaviors during fusion.

Several studies illustrate fused safety risks. \citet{WalmerSSSJ22} embed backdoor triggers in both image and text, where each modality appears benign in isolation but induces malicious behavior when combined. \citet{abs-2402-08577} show asymmetric modality roles, with images facilitating trigger injection and text activating malicious responses. 
\citet{ImgTrojan} associate harmful queries with specific clean images by replacing captions with jailbreak prompts, causing model generates harmful content when presented with both.
\citet{FigStep} embed jailbreak prompts in visual inputs and use benign-looking text to elicit harmful outputs.
\citet{song2025jailbound} approximate safety decision boundaries in multi-modal fusion spaces using logistic regression and manipulate fusion-layer features toward a target decision to induce policy-violating outputs.

Beyond attacks, modality fusion can also induce hallucinations when cross-modal evidence is imbalanced, allowing dominant linguistic priors to override weaker visual or auditory signals. This effect is empirically shown in Curse of Multi-Modalities \citep{leng2024}, where MLLMs frequently default to unimodal dominance, leading to systematic hallucinations under fusion conflicts.


\vspace{-0.1in}
\subsection{Shifts in Threat Models}
\vspace{-0.1in}

Based on the above discussion of emerging threats, we summarize the key shifts in threat modeling introduced by multi-modal settings compared to uni-modal models.
(1) \textbf{Relaxed Capability:} 
In multi-modal systems, adversaries no longer need full-system access. Due to the compositional design, compromising a subset of modalities can suffice. For example, adversarial images targeting only the visual encoder can trigger downstream misalignment during fusion, resulting in harmful or misleading outputs.
(2) \textbf{Expanded Threat Surface:} 
Multi-modal systems introduce new threat surfaces beyond input–output mapping, including vulnerabilities in modality alignment and fusion.
These internal stages create deeper, stealthier entry points for threats within the model pipeline.
(3) \textbf{Cross-Modality Threats:}  
Multi-modal models enable threats that exploit interactions across modalities. For example, an adversarial image may trigger a jailbreak, or a malicious prompt may mislead the model’s interpretation of visual input.

\vspace{-0.1in}
\section{Current Multi-modal Safety Solutions}\label{sec:solutions}
\vspace{-0.1in}

In this section, we summarize new safety constraints in multi-modal systems driven by evolving threat models and review recent advances in safety fine-tuning, preference-based optimization, and training-free approaches.

\begin{table*}[t]
    \centering
    \vspace{-0.2in}
    \resizebox{0.98\linewidth}{!}{
    \begin{tabular}{c|c|c|c|c|c|c}
    \toprule
        \multirow{2}{*}{\textbf{Category}} & \multirow{2}{*}{\textbf{Sub-category}} & \multirow{2}{*}{\textbf{Paper}} & \multicolumn{3}{c|}{\textbf{Safety Constraints}} & \multirow{2}{*}{\textbf{Year \& Venue}} \\
        \cline{4-6}
        & & & \textbf{Cap. 1} & \textbf{Cap. 2} & \textbf{Cap. 3} & \\
        \hline
        \multirow{11}{*}{\textbf{\shortstack{Safety \\ Fine-Tuning}}} 
        & \multirow{2}{*}{\shortstack{Modality-Wise \\ Fine-Tuning}}
            & Robust CLIP\cite{SchlarmannSC024} & & & \checkmark &  ICML'24 \\
            & & Textual Unlearning\cite{ChakrabortySCAA24} & & & \checkmark & EMNLP'24 Findings  \\
        \cline{2-7}
        & \multirow{9}{*}{\shortstack{Cross-Modal \\ Fine-Tuning}} 
            & TGA-ZSR \cite{0004ZX24} & & \checkmark & \checkmark & NeurIPS'24 \\
            & & VLGuard\cite{Safety_FT} & \checkmark & \checkmark & & ICML'24 \\
            & & Hallusionbench\cite{guan2024hallusionbench} & \checkmark & \checkmark & & CVPR'24 \\
            & & Mmc\cite{liu2024mmc} & \checkmark & \checkmark & & NAACL'24 \\
            & & Robust-VLGuard\cite{Safeguard_VLM} & & \checkmark & & ICCV'25 \\
            & & TGA \cite{XuPZSC25}  & & \checkmark & \checkmark  & ICLR'25 \\
            & & BYE \cite{BYE} & \checkmark & \checkmark & & NeurIPS'25 \\
            & & UNIGUARD\cite{UniGuard} & \checkmark & & & arXiv'25 \\
            & & ProEAT\cite{lu2025adversarialtrainingmultimodallarge} & \checkmark & \checkmark & & arXiv'25 \\
        \hline
        \multirow{6}{*}{\textbf{\shortstack{Preference-Based \\ Optimization}}} & \multirow{4}{*}{Human Feedback}
        & SPA-VL\cite{ZhangCZGZFYJQ0Z25} & \checkmark & \checkmark & & CVPR'25 \\
        & & AdPO\cite{abs-2504-01735} & & \checkmark & \checkmark & arXiv’25 \\
        & & GuardReasoner-VL\cite{liu2025guardreasonervl} & \checkmark & \checkmark & & NeurIPS'25 \\
        & & Adversary-aware DPO\cite{Weng25} & \checkmark & \checkmark & \checkmark & EMNLP'25 Findings \\
        \cline{2-7}
        & \multirow{2}{*}{AI Feedback}
        & Silkie\cite{li2023silkie} & \checkmark & \checkmark & & arXiv'23  \\
        & & MHALO\cite{cai2025mhalo} & \checkmark & \checkmark & & ACL'25 Findings \\
        \hline
        \multirow{15}{*}{\textbf{\shortstack{Training-Free \\ Solutions}}} & \multirow{3}{*}{\shortstack{Input-level \\ Safeguard}}
        & AdaShield~\cite{WangLLCX24} & \checkmark & \checkmark & &  ECCV'24 \\
        & & BlueSuffix~\cite{ZhaoZLLM025}  & \checkmark & \checkmark &  & ICLR'25 \\
        & & LLM-Pipeline\cite{guo2025} & & \checkmark & & NeurIPS'25 \\
        \cline{2-7}
        & \multirow{8}{*}{\shortstack{Internal-level \\ Intervention}}
        & InferAligner\cite{WangZLTWZRJQ24} & & \checkmark & \checkmark & EMNLP'24 \\
        & & ASTRA \cite{WangWZ25} & & \checkmark & \checkmark & CVPR'25 \\
        & & ETA\cite{DingLZ25}  &  & \checkmark & \checkmark & ICLR'25 \\
        & & HiddenDetect\cite{hiddendetect} & \checkmark & \checkmark & \checkmark & ACL'25 \\
        & & SafePTR\cite{chen2025safeptr} & \checkmark & \checkmark & \checkmark & NeurIPS'25 \\
        & & ALFAR\cite{anboosting} & \checkmark & \checkmark & \checkmark & NeurIPS'25 \\
        & & SASA\cite{SASA} & \checkmark & \checkmark & \checkmark & ACM MM'25 \\
        & & CIDER\cite{XuQQW24} & & \checkmark & \checkmark & EMNLP'25 Findings \\
        \cline{2-7}
        & \multirow{4}{*}{\shortstack{Output-level \\ Control}}
        & MLLM-Protector\cite{PiHZXPLDZZ24} & \checkmark & \checkmark & & EMNLP'24 \\
        & & ECSO\cite{ECSO} & & \checkmark & & ECCV'24 \\
        & & IMMUNE\cite{Ghosal_2025_CVPR} & \checkmark & \checkmark & \checkmark & CVPR'25 \\
        & & DeGF\cite{zhang2025self} &  & \checkmark & & ICLR'25 \\
    \bottomrule
    \end{tabular}
    }
    \vspace{-0.05in}
    \caption{Recent advancements in multi-modal safety solutions. Cap. 1: Modality-Agnostic Readiness; Cap. 2: Cross-Modal Safety; Cap. 3: Internal Safety Intervention.}
    \label{tab:solution}
    \vspace{-0.2in}
\end{table*}


\vspace{-0.1in}
\subsection{New Safety Constraints}
\vspace{-0.1in}
(1) \textbf{Modality-Agnostic Readiness:}
Multi-modal systems face dynamic threats across modalities, making fixed assumptions about threat types or entry modalities infeasible.
For example, relying solely on text-based safety mechanisms is insufficient -- adversarial visual inputs can bypass language-focused safeguards by embedding hidden jailbreak triggers, exposing critical blind spots in classic threat assumptions.
(2) \textbf{Cross-Modal Safety:}
Designing and ensembling modality-isolated safety mechanisms can disrupt cross-modal interactions and introduce unintended inconsistencies. Instead, safety solutions should operate holistically across modalities to preserve semantic coherence and support effective alignment and fusion. 
(3) \textbf{Internal Safety Intervention:} 
Traditional safety mechanisms typically focus on monitoring model inputs and outputs. However, in multi-modal systems, threats can emerge at intermediate stages --modality alignment and fusion. To address such risks, safety solutions should intervene at internal stages, rather than relying solely on surface-level observations.

\vspace{-0.1in}
\subsection{Safety Fine-Tuning}
\vspace{-0.1in}

Safety fine-tuning adapts pre-trained models with safety-oriented data without full retraining, including modality-wise fine-tuning within individual modalities or components  (e.g., vision encoders or LLMs) and cross-modal fine-tuning that aligns interactions across modalities.

Several studies adopt modality-wise fine-tuning as a cost-effective safety strategy. 
On the vision side, works such as~\cite{SchlarmannSC024,MaoGYWV23,001200S24,shi2024eagle} improve robustness by fine-tuning the visual encoder to resist adversarial perturbations while preserving feature representations.
\citet{0004ZX24} further leverage text embeddings to guide attention and correct adversarial visual shifts.
In contrast, other approaches enhance overall VLM safety by intervening in the LLM component. For example, \citet{ChakrabortySCAA24} apply textual unlearning to prevent harmful context from propagating in the LLM’s latent space and redirect it toward safer representations.

Other works investigate cross-modal fine-tuning with safety-aligned datasets.
VLGuard~\citep{Safety_FT} and Robust-VLGuard~\citep{Safeguard_VLM} provide safety-aligned datasets with both aligned and misaligned image-text pairs, improving MLLMs resilience against Gaussian noise and jailbreaks.
UNIGUARD~\citep{UniGuard} models both unimodal and cross-modal harmful signals and enforces interaction-aware defenses across modalities.
ProEAT~\citep{lu2025adversarialtrainingmultimodallarge} applies adversarial training to both visual projection layers and LLM to enhance cross-modal consistency and robustness.
BYE~\citep{BYE} filters poisoned fine-tuning samples by detecting anomalous low-entropy cross-modal attention patterns.
\citet{XuPZSC25} guide the projection of visual embeddings during alignment using retrieved textual evidence to mitigate toxic-image-induced risks.

Beyond attack scenarios, recent studies~\cite{guan2024hallusionbench, li2025benchmark} show that hallucination in MLLMs is also strongly influenced by instruction-tuning data.
LRV-Instruction \cite{liu2023mitigating,liu2024mmc} reduces instruction bias by incorporating both positive and negative instruction samples. 
PerturboLLaVA~\cite{chen2025perturbollava} introduces adversarial instruction-vision conflicts to discourage over-reliance on language priors. 
REVERIE~\cite{fei2024multimodal} further incorporates reasoning supervision by requiring rationales for correct and incorrect responses, promoting more robust cross-modal reasoning.

\vspace{-0.1in}
\subsection{Preference-Based Optimization}
\vspace{-0.1in}

Preference-based optimization improves safety alignment by optimizing over relative preferences between model outputs, rather than explicit labels or losses, using human or AI feedback to favor preferred responses over rejected ones.

Recent studies adopt preference-based optimization to align MLLMs under adversarial settings. 
SPA-VL~\citep{ZhangCZGZFYJQ0Z25} introduces a large-scale safety alignment dataset with preferred and rejected responses for supervised fine-tuning and subsequent RLHF/Direct Preference Optimization (DPO) training.
GuardReasoner-VL~\citep{liu2025guardreasonervl} adopts a reason-then-moderate paradigm with online RL to detect hidden harmful content.
\citet{abs-2504-01735} combine classic adversarial training with DPO to update the visual encoder against adversarial image attacks, improving multi-modal safety alignment. 
Similarly, \citet{Weng25} propose an adversary-aware DPO framework, in which the reference model is adversarially trained to provide reliable supervision, and a min-max objective is applied with perturbations in both image and latent spaces.

Beyond human feedback, several studies explore RL from AI feedback for hallucination mitigation. 
HA-DPO~\cite{zhao2023beyond} and Silkie~\cite{li2023silkie} apply DPO to favor accurate over hallucinated responses using AI-generated feedback or stronger teacher model preferences.
Challenging standard DPO assumptions, POViD~\cite{li2025self} constructs preference data solely from hallucinated responses, treating ground-truth instructions as preferred outputs, while HalDetect~\cite{cai2025mhalo} introduces fine-grained hallucination-aware rewards and FDPO to leverage human feedback at finer granularity.

\vspace{-0.1in}
\subsection{Training-Free Solutions}
\vspace{-0.1in}

Training-free solutions improve MLLM safety at inference time without updating model parameters, operating at the input level to block harmful or adversarial content, the internal level to suppress unsafe behaviors via intermediate representations, and the output level to verify or revise generations for safer responses.

At the input level, \citet{guo2025} use an auxiliary LLM as a vision-free filter to screen and block harmful prompts before they are passed to the multi-modal model. 
AdaShield~\cite{WangLLCX24} prepends model inputs with input-aware defense prompts that automatically and adaptively safeguard MLLMs against jailbreak attacks.
DiffPure-VLM~\cite{Safeguard_VLM} and BlueSuffix~\cite{ZhaoZLLM025} mitigate visual adversarial perturbations using diffusion-based models.
BlueSuffix further improves safety by rewriting textual prompts with an LLM-based purifier and prepending LLM-generated suffixes to the input.

For intervention in internal layers, CIDER~\cite{XuQQW24} detects image-induced jailbreaks by identifying inputs with abnormally low cross-modal semantic distances in the embedding space.
InferAligner~\cite{WangZLTWZRJQ24} and ASTRA~\cite{WangWZ25} suppress jailbreak-related hidden representations using steering vectors, with InferAligner deriving safety steering vectors from aligned LLMs and ASTRA extracting adversarial steering vectors via image-based attribution.
SASA~\cite{SASA} and ETA~\cite{DingLZ25} adopt multi-stage safety mechanisms. SASA projects semantically rich representations from deeper fused layers to early layers to enhance risk perception, while ETA combines shallow alignment with deep sentence-level selection to enforce safety.
SafePTR~\cite{chen2025safeptr} and HiddenDetect~\cite{hiddendetect} partition models into vulnerable and safety-relevant layers. SafePTR prunes harmful cross-modal tokens in vulnerable layers, whereas HiddenDetect detects unsafe prompts by measuring the alignment of activations with a refusal-aware embedding across safety layers.

For output-level control, \citet{PiHZXPLDZZ24} assess response harmfulness using a harm detector and revise unsafe outputs with a detoxifier.
IMMUNE~\cite{Ghosal_2025_CVPR} guides token selection during decoding using a learned safety reward model to favor safer continuations. 
\citet{ECSO} enable self-checking in MLLMs and regenerate safer outputs by rerouting visual inputs through an alignment-protected text-only pipeline.

Multi-modal hallucination can be mitigated by guided or self-correcting decoding.
DeGF~\cite{zhang2025self} verifies and corrects model responses using generative feedback from text-to-image models, while MARINE~\cite{zhao2024mitigating} guides decoding with grounded object representations and classifier-free guidance to emphasize visual evidence. 
At the attention level, EAH~\cite{li2025videohalluevaluatingmitigatingmultimodal} and VHR~\cite{zhang2024seeing} mitigate visual attention sink by strengthening image-token attention in shallow layers, while \citet{anboosting} adaptively rebalances attention in multi-modal RAG based on retrieval scores and query-context relevance. 
To reduce over-reliance on language priors, Woodpecker~\cite{yin2024woodpecker} detects and corrects hallucinations using visual experts, and \citet{liu2025hallucinations} encourage greater reliance on visual context.

\vspace{-0.1in}
\section{Future Directions}\label{sec:future}
\vspace{-0.1in}


Despite progress, MLLM safety remains challenged by expanded attack surfaces, cross-modal interactions, and new constraints, motivating future research.

\vspace{-0.1in}
\subsection{Resilience to Partial Corruption}
\vspace{-0.1in}

As discussed in Section~\ref{sec:threats}, partial corruption can undermine system safety. Although MLLMs face expanded attack surfaces and modality inconsistencies, they also benefit from overlapping cross-modal information that can support robustness when some modalities are compromised. To avoid new points of failure, models should not over-rely on any subset of modalities. Future defenses should instead exploit cross-modal redundancy through mechanisms such as selective modality rejection, confidence-aware fusion, or adaptive weighting, enabling models to down-weight unreliable modalities while preserving performance and robustness.


\vspace{-0.1in}
\subsection{Toward Unified Safety Framework}
\vspace{-0.1in}

Many existing methods apply modality-specific defenses independently, but such ensemble-style strategies are neither scalable nor well aligned with multi-modal principles. Since not all modalities are compromised simultaneously, uniform defenses are often unnecessary and inefficient, while aggressive filtering can disrupt cross-modal coherence and degrade alignment and fusion. Future work should pursue unified, modality- and threat-agnostic safety frameworks, particularly by extending defenses to shared embedding and fusion stages and explicitly accounting for modality alignment through alignment-aware training or cross-modal consistency regularization.


\vspace{-0.1in}
\subsection{Hallucination Beyond Text}
\vspace{-0.1in}

Hallucination remains a fundamental risk in MLLMs, and current approaches lack principled and scalable solutions. Future work should improve data quality and calibration across modalities, strengthen cross-modal alignment through enhanced representations and consistency regularization, and pursue architectural advances that mitigate hallucination at its source. Progress also depends on theoretically grounded, human-aligned, and truly multimodal evaluation benchmarks beyond simplified or model-based assessments.


\vspace{-0.1in}
\subsection{From Manual to Agent-Driven Safety}
\vspace{-0.1in}

Existing AI safety frameworks largely rely on manual efforts to identify vulnerabilities and design safeguards, an approach that is increasingly insufficient for modern multi-modal systems. As discussed in Sections~\ref{sec:threats} and~\ref{sec:solutions}, the threat surface of MLLMs has expanded into a high-dimensional and dynamic space spanning multiple modalities, cross-modal interactions, and diverse attack patterns, making systematic manual exploration infeasible. Moreover, many defenses are evaluated under static threat assumptions or limited attack settings without considering adaptive adversaries~\citep{attackermovessecond}. To address these challenges, we advocate a shift toward automated, agent-driven red-blue teaming, where red-teaming agents autonomously explore complex attack strategies and discovered vulnerabilities guide blue-teaming mechanisms for the continuous development of more robust and proactive safety solutions.

\vspace{-0.1in}
\section{Conclusion}\label{sec:conclusion}
\vspace{-0.1in}

In response to the evolving safety landscape, this survey introduces a new taxonomy of multi-modal safety risks -- compromised modality integration, modality misalignment, and fused safety risks -- and highlights shifts in threat assumptions under compositional architectures. We further summarize emerging safety constraints and recent advances in safety solutions, including safety fine-tuning, preference-based optimization, and training-free approaches, and outline future directions toward unified and robust safety frameworks for next-generation multi-modal AI systems.


\clearpage
\newpage
\bibliography{iclr2026_conference}

@inproceedings{LLAVA,
author = {Liu, Haotian and Li, Chunyuan and Wu, Qingyang and Lee, Yong Jae},
title = {Visual instruction tuning},
year = {2023},
booktitle = {Proceedings of the 37th International Conference on Neural Information Processing Systems},
articleno = {1516},
numpages = {25},
location = {New Orleans, LA, USA},
series = {NIPS '23}
}

@misc{leng2024,
      title={The Curse of Multi-Modalities: Evaluating Hallucinations of Large Multimodal Models across Language, Visual, and Audio}, 
      author={Sicong Leng and Yun Xing and Zesen Cheng and Yang Zhou and Hang Zhang and Xin Li and Deli Zhao and Shijian Lu and Chunyan Miao and Lidong Bing},
      year={2024},
      eprint={2410.12787},
      archivePrefix={arXiv},
      primaryClass={cs.CV},
      url={https://arxiv.org/abs/2410.12787}, 
}

@article{li2025self,
  title={Self-Rewarding Vision-Language Model via Reasoning Decomposition},
  author={Li, Zongxia and Yu, Wenhao and Huang, Chengsong and Liu, Rui and Liang, Zhenwen and Liu, Fuxiao and Che, Jingxi and Yu, Dian and Boyd-Graber, Jordan and Mi, Haitao and others},
  journal={arXiv preprint arXiv:2508.19652},
  year={2025}
}

@inproceedings{gao2023pal,
  title={Pal: Program-aided language models},
  author={Gao, Luyu and Madaan, Aman and Zhou, Shuyan and Alon, Uri and Liu, Pengfei and Yang, Yiming and Callan, Jamie and Neubig, Graham},
  booktitle={ICML},
  pages={10764--10799},
  year={2023},
  organization={PMLR}
}

@article{izacard2020leveraging,
  title={Leveraging passage retrieval with generative models for open domain question answering},
  author={Izacard, Gautier and Grave, Edouard},
  journal={arXiv preprint arXiv:2007.01282},
  year={2020}
}

@misc{liu2024llavanext,
    title={LLaVA-NeXT: Improved reasoning, OCR, and world knowledge},
    url={https://llava-vl.github.io/blog/2024-01-30-llava-next/},
    author={Liu, Haotian and Li, Chunyuan and Li, Yuheng and Li, Bo and Zhang, Yuanhan and Shen, Sheng and Lee, Yong Jae},
    month={January},
    year={2024}
}

@inproceedings{ye2024mplug,
  title={mplug-owl2: Revolutionizing multi-modal large language model with modality collaboration},
  author={Ye, Qinghao and Xu, Haiyang and Ye, Jiabo and Yan, Ming and Hu, Anwen and Liu, Haowei and Qian, Qi and Zhang, Ji and Huang, Fei},
  booktitle={Proceedings of the ieee/cvf conference on computer vision and pattern recognition},
  pages={13040--13051},
  year={2024}
}

@article{zhu2023minigpt,
  title={Minigpt-4: Enhancing vision-language understanding with advanced large language models},
  author={Zhu, Deyao and Chen, Jun and Shen, Xiaoqian and Li, Xiang and Elhoseiny, Mohamed},
  journal={arXiv preprint arXiv:2304.10592},
  year={2023}
}

@misc{dai2023instructblipgeneralpurposevisionlanguagemodels,
      title={InstructBLIP: Towards General-purpose Vision-Language Models with Instruction Tuning}, 
      author={Wenliang Dai and Junnan Li and Dongxu Li and Anthony Meng Huat Tiong and Junqi Zhao and Weisheng Wang and Boyang Li and Pascale Fung and Steven Hoi},
      year={2023},
      eprint={2305.06500},
      archivePrefix={arXiv},
      primaryClass={cs.CV},
      url={https://arxiv.org/abs/2305.06500}, 
}

@inproceedings{li2023blip,
  title={Blip-2: Bootstrapping language-image pre-training with frozen image encoders and large language models},
  author={Li, Junnan and Li, Dongxu and Savarese, Silvio and Hoi, Steven},
  booktitle={International conference on machine learning},
  pages={19730--19742},
  year={2023},
  organization={PMLR}
}

@inproceedings{karamcheti2024prismatic,
  title={Prismatic vlms: Investigating the design space of visually-conditioned language models},
  author={Karamcheti, Siddharth and Nair, Suraj and Balakrishna, Ashwin and Liang, Percy and Kollar, Thomas and Sadigh, Dorsa},
  booktitle={Forty-first International Conference on Machine Learning},
  year={2024}
}

@article{shi2024eagle,
  title={Eagle: Exploring the design space for multimodal llms with mixture of encoders},
  author={Shi, Min and Liu, Fuxiao and Wang, Shihao and Liao, Shijia and Radhakrishnan, Subhashree and Zhao, Yilin and Huang, De-An and Yin, Hongxu and Sapra, Karan and Yacoob, Yaser and others},
  journal={arXiv preprint arXiv:2408.15998},
  year={2024}
}

@article{beurer2023prompting,
  title={Prompting is programming: A query language for large language models},
  author={Beurer-Kellner, Luca and Fischer, Marc and Vechev, Martin},
  journal={Proceedings of the ACM on Programming Languages},
  volume={7},
  number={PLDI},
  pages={1946--1969},
  year={2023},
  publisher={ACM New York, NY, USA}
}

@inproceedings{li2023making,
  title={Making language models better reasoners with step-aware verifier},
  author={Li, Yifei and Lin, Zeqi and Zhang, Shizhuo and Fu, Qiang and Chen, Bei and Lou, Jian-Guang and Chen, Weizhu},
  booktitle={ACL},
  year={2023}
}

@article{zhou2022least,
  title={Least-to-most prompting enables complex reasoning in large language models},
  author={Zhou, Denny and Sch{\"a}rli, Nathanael and Hou, Le and Wei, Jason and Scales, Nathan and Wang, Xuezhi and Schuurmans, Dale and Cui, Claire and Bousquet, Olivier and Le, Quoc and others},
  journal={arXiv preprint arXiv:2205.10625},
  year={2022}
}

@article{lewis2020retrieval,
  title={Retrieval-augmented generation for knowledge-intensive nlp tasks},
  author={Lewis, Patrick and Perez, Ethan and Piktus, Aleksandra and Petroni, Fabio and Karpukhin, Vladimir and Goyal, Naman and K{\"u}ttler, Heinrich and Lewis, Mike and Yih, Wen-tau and Rockt{\"a}schel, Tim and others},
  journal={Advances in neural information processing systems},
  volume={33},
  pages={9459--9474},
  year={2020}
}

@article{ziegler2019fine,
  title={Fine-tuning language models from human preferences},
  author={Ziegler, Daniel M and Stiennon, Nisan and Wu, Jeffrey and Brown, Tom B and Radford, Alec and Amodei, Dario and Christiano, Paul and Irving, Geoffrey},
  journal={arXiv preprint arXiv:1909.08593},
  year={2019}
}

@article{wei2022chain,
  title={Chain-of-thought prompting elicits reasoning in large language models},
  author={Wei, Jason and Wang, Xuezhi and Schuurmans, Dale and Bosma, Maarten and Xia, Fei and Chi, Ed and Le, Quoc V and Zhou, Denny and others},
  journal={Advances in neural information processing systems},
  volume={35},
  pages={24824--24837},
  year={2022}
}

@article{liu2023mitigating,
  title={Mitigating hallucination in large multi-modal models via robust instruction tuning},
  author={Liu, Fuxiao and Lin, Kevin and Li, Linjie and Wang, Jianfeng and Yacoob, Yaser and Wang, Lijuan},
  journal={arXiv preprint arXiv:2306.14565},
  year={2023}
}

@inproceedings{guan2024hallusionbench,
  title={Hallusionbench: an advanced diagnostic suite for entangled language hallucination and visual illusion in large vision-language models},
  author={Guan, Tianrui and Liu, Fuxiao and Wu, Xiyang and Xian, Ruiqi and Li, Zongxia and Liu, Xiaoyu and Wang, Xijun and Chen, Lichang and Huang, Furong and Yacoob, Yaser and others},
  booktitle={CVPR},
  year={2024}
}

@inproceedings{liu2024mmc,
  title={Mmc: Advancing multimodal chart understanding with large-scale instruction tuning},
  author={Liu, Fuxiao and Wang, Xiaoyang and Yao, Wenlin and Chen, Jianshu and Song, Kaiqiang and Cho, Sangwoo and Yacoob, Yaser and Yu, Dong},
  booktitle={NAACL},
  year={2024}
}

@phdthesis{liu2025hallucinations,
  title={Hallucinations in Multimodal Large Language Models: Evaluation, Mitigation, and Future Directions},
  author={Liu, Fuxiao},
  year={2025},
  school={University of Maryland, College Park}
}

@misc{li2025videohalluevaluatingmitigatingmultimodal,
      title={VideoHallu: Evaluating and Mitigating Multi-modal Hallucinations on Synthetic Video Understanding}, 
      author={Zongxia Li and Xiyang Wu and Guangyao Shi and Yubin Qin and Hongyang Du and Fuxiao Liu and Tianyi Zhou and Dinesh Manocha and Jordan Lee Boyd-Graber},
      year={2025},
      eprint={2505.01481},
      archivePrefix={arXiv},
      primaryClass={cs.CV},
      url={https://arxiv.org/abs/2505.01481}, 
}

@article{yin2024woodpecker,
  title={Woodpecker: Hallucination correction for multimodal large language models},
  author={Yin, Shukang and Fu, Chaoyou and Zhao, Sirui and Xu, Tong and Wang, Hao and Sui, Dianbo and Shen, Yunhang and Li, Ke and Sun, Xing and Chen, Enhong},
  journal={Science China Information Sciences},
  volume={67},
  number={12},
  pages={220105},
  year={2024},
  publisher={Springer}
}

@article{zhang2024seeing,
  title={Seeing clearly by layer two: Enhancing attention heads to alleviate hallucination in lvlms},
  author={Zhang, Xiaofeng and Quan, Yihao and Gu, Chaochen and Shen, Chen and Yuan, Xiaosong and Yan, Shaotian and Cheng, Hao and Wu, Kaijie and Ye, Jieping},
  journal={arXiv preprint arXiv:2411.09968},
  year={2024}
}

@article{zhao2024mitigating,
  title={Mitigating object hallucination in large vision-language models via classifier-free guidance},
  author={Zhao, Linxi and Deng, Yihe and Zhang, Weitong and Gu, Quanquan},
  journal={arXiv e-prints},
  pages={arXiv--2402},
  year={2024}
}

@inproceedings{zhang2025self,
  author       = {Ce Zhang and
                  Zifu Wan and
                  Zhehan Kan and
                  Martin Q. Ma and
                  Simon Stepputtis and
                  Deva Ramanan and
                  Russ Salakhutdinov and
                  Louis{-}Philippe Morency and
                  Katia P. Sycara and
                  Yaqi Xie},
  title        = {Self-Correcting Decoding with Generative Feedback for Mitigating Hallucinations
                  in Large Vision-Language Models},
  booktitle    = {{ICLR}},
  year         = {2025},
}

@article{li2025benchmark,
  title={Benchmark evaluations, applications, and challenges of large vision language models: A survey},
  author={Li, Zongxia and Wu, Xiyang and Du, Hongyang and Nghiem, Huy and Shi, Guangyao},
  journal={arXiv preprint arXiv:2501.02189},
  volume={1},
  year={2025}
}

@article{li2023silkie,
  title={Silkie: Preference distillation for large visual language models},
  author={Li, Lei and Xie, Zhihui and Li, Mukai and Chen, Shunian and Wang, Peiyi and Chen, Liang and Yang, Yazheng and Wang, Benyou and Kong, Lingpeng},
  journal={arXiv preprint arXiv:2312.10665},
  year={2023}
}

@article{zhao2023beyond,
  title={Beyond hallucinations: Enhancing lvlms through hallucination-aware direct preference optimization},
  author={Zhao, Zhiyuan and Wang, Bin and Ouyang, Linke and Dong, Xiaoyi and Wang, Jiaqi and He, Conghui},
  journal={arXiv preprint arXiv:2311.16839},
  year={2023}
}

@inproceedings{cai2025mhalo,
  title={MHALO: Evaluating MLLMs as Fine-grained Hallucination Detectors},
  author={Cai, Yishuo and Gu, Renjie and Li, Jiaxu and Huang, Xuancheng and Chen, Junzhe and Gu, Xiaotao and Huang, Minlie},
  booktitle={Findings of ACL},
  year={2025}
}

@article{chen2025perturbollava,
  title={PerturboLLaVA: Reducing multimodal hallucinations with perturbative visual training},
  author={Chen, Cong and Liu, Mingyu and Jing, Chenchen and Zhou, Yizhou and Rao, Fengyun and Chen, Hao and Zhang, Bo and Shen, Chunhua},
  journal={arXiv preprint arXiv:2503.06486},
  year={2025}
}

@inproceedings{fei2024multimodal,
  title={From multimodal llm to human-level ai: Modality, instruction, reasoning, efficiency and beyond},
  author={Fei, Hao and Yao, Yuan and Zhang, Zhuosheng and Liu, Fuxiao and Zhang, Ao and Chua, Tat-Seng},
  booktitle={Proceedings of the 2024 Joint International Conference on Computational Linguistics, Language Resources and Evaluation (LREC-COLING 2024): Tutorial Summaries},
  pages={1--8},
  year={2024}
}

@inproceedings{sun2023contrastive,
  title={Contrastive learning reduces hallucination in conversations},
  author={Sun, Weiwei and Shi, Zhengliang and Gao, Shen and Ren, Pengjie and de Rijke, Maarten and Ren, Zhaochun},
  booktitle={AAAI},
  year={2023}
}

@article{raffel2020exploring,
  title={Exploring the limits of transfer learning with a unified text-to-text transformer},
  author={Raffel, Colin and Shazeer, Noam and Roberts, Adam and Lee, Katherine and Narang, Sharan and Matena, Michael and Zhou, Yanqi and Li, Wei and Liu, Peter J},
  journal={Journal of machine learning research},
  volume={21},
  number={140},
  pages={1--67},
  year={2020}
}

@article{huang2025survey,
  title={A survey on hallucination in large language models: Principles, taxonomy, challenges, and open questions},
  author={Huang, Lei and Yu, Weijiang and Ma, Weitao and Zhong, Weihong and Feng, Zhangyin and Wang, Haotian and Chen, Qianglong and Peng, Weihua and Feng, Xiaocheng and Qin, Bing and others},
  journal={ACM Transactions on Information Systems},
  volume={43},
  number={2},
  pages={1--55},
  year={2025},
  publisher={ACM New York, NY}
}

@inproceedings{bender2021dangers,
  title={On the dangers of stochastic parrots: Can language models be too big?},
  author={Bender, Emily M and Gebru, Timnit and McMillan-Major, Angelina and Shmitchell, Shmargaret},
  booktitle={Proceedings of the 2021 ACM conference on fairness, accountability, and transparency},
  pages={610--623},
  year={2021}
}

@article{sheng2020towards,
  title={Towards controllable biases in language generation},
  author={Sheng, Emily and Chang, Kai-Wei and Natarajan, Premkumar and Peng, Nanyun},
  journal={arXiv preprint arXiv:2005.00268},
  year={2020}
}

@inproceedings{guu2020retrieval,
  title={Retrieval augmented language model pre-training},
  author={Guu, Kelvin and Lee, Kenton and Tung, Zora and Pasupat, Panupong and Chang, Mingwei},
  booktitle={ICML},
  pages={3929--3938},
  year={2020},
  organization={PMLR}
}

@article{penedo2023refinedweb,
  title={The RefinedWeb dataset for Falcon LLM: outperforming curated corpora with web data, and web data only},
  author={Penedo, Guilherme and Malartic, Quentin and Hesslow, Daniel and Cojocaru, Ruxandra and Cappelli, Alessandro and Alobeidli, Hamza and Pannier, Baptiste and Almazrouei, Ebtesam and Launay, Julien},
  journal={arXiv preprint arXiv:2306.01116},
  year={2023}
}

@article{lee2021deduplicating,
  title={Deduplicating training data makes language models better},
  author={Lee, Katherine and Ippolito, Daphne and Nystrom, Andrew and Zhang, Chiyuan and Eck, Douglas and Callison-Burch, Chris and Carlini, Nicholas},
  journal={arXiv preprint arXiv:2107.06499},
  year={2021}
}

@article{farquhar2024detecting,
  title={Detecting hallucinations in large language models using semantic entropy},
  author={Farquhar, Sebastian and Kossen, Jannik and Kuhn, Lorenz and Gal, Yarin},
  journal={Nature},
  volume={630},
  number={8017},
  pages={625--630},
  year={2024},
  publisher={Nature Publishing Group UK London}
}

@article{tu2020empirical,
  title={An empirical study on robustness to spurious correlations using pre-trained language models},
  author={Tu, Lifu and Lalwani, Garima and Gella, Spandana and He, He},
  journal={Transactions of the Association for Computational Linguistics},
  volume={8},
  pages={621--633},
  year={2020},
  publisher={MIT Press One Rogers Street, Cambridge, MA 02142-1209, USA journals-info~…}
}

@article{wallace2019universal,
  title={Universal adversarial triggers for attacking and analyzing NLP},
  author={Wallace, Eric and Feng, Shi and Kandpal, Nikhil and Gardner, Matt and Singh, Sameer},
  journal={arXiv preprint arXiv:1908.07125},
  year={2019}
}

@article{wang2021identifying,
  title={Identifying and mitigating spurious correlations for improving robustness in nlp models},
  author={Wang, Tianlu and Sridhar, Rohit and Yang, Diyi and Wang, Xuezhi},
  journal={arXiv preprint arXiv:2110.07736},
  year={2021}
}

@article{mckenna2023sources,
  title={Sources of hallucination by large language models on inference tasks},
  author={McKenna, Nick and Li, Tianyi and Cheng, Liang and Hosseini, Mohammad Javad and Johnson, Mark and Steedman, Mark},
  journal={arXiv preprint arXiv:2305.14552},
  year={2023}
}

@inproceedings{Flamingo,
author = {Alayrac, Jean-Baptiste and Donahue, Jeff and Luc, Pauline and Miech, Antoine and Barr, Iain and Hasson, Yana and Lenc, Karel and Mensch, Arthur and Millicah, Katie and Reynolds, Malcolm and Ring, Roman and Rutherford, Eliza and Cabi, Serkan and Han, Tengda and Gong, Zhitao and Samangooei, Sina and Monteiro, Marianne and Menick, Jacob and Borgeaud, Sebastian and Brock, Andrew and Nematzadeh, Aida and Sharifzadeh, Sahand and Binkowski, Mikolaj and Barreira, Ricardo and Vinyals, Oriol and Zisserman, Andrew and Simonyan, Karen},
title = {Flamingo: a visual language model for few-shot learning},
year = {2022},
isbn = {9781713871088},
booktitle = {Proceedings of the 36th International Conference on Neural Information Processing Systems},
articleno = {1723},
numpages = {21},
location = {New Orleans, LA, USA},
series = {NIPS '22}
}

@inproceedings{BLIP,
title={BLIP: Bootstrapping Language-Image Pre-training for Unified Vision-Language Understanding and Generation}, 
author={Junnan Li and Dongxu Li and Caiming Xiong and Steven Hoi},
year={2022},
booktitle={ICML},
}

@inproceedings{GPT,
author = {Brown, Tom B. and Mann, Benjamin and Ryder, Nick and Subbiah, Melanie and Kaplan, Jared and Dhariwal, Prafulla and Neelakantan, Arvind and Shyam, Pranav and Sastry, Girish and Askell, Amanda and Agarwal, Sandhini and Herbert-Voss, Ariel and Krueger, Gretchen and Henighan, Tom and Child, Rewon and Ramesh, Aditya and Ziegler, Daniel M. and Wu, Jeffrey and Winter, Clemens and Hesse, Christopher and Chen, Mark and Sigler, Eric and Litwin, Mateusz and Gray, Scott and Chess, Benjamin and Clark, Jack and Berner, Christopher and McCandlish, Sam and Radford, Alec and Sutskever, Ilya and Amodei, Dario},
title = {Language models are few-shot learners},
year = {2020},
isbn = {9781713829546},
booktitle = {Proceedings of the 34th International Conference on Neural Information Processing Systems},
articleno = {159},
numpages = {25},
location = {Vancouver, BC, Canada},
series = {NIPS '20}
}

@misc{DeepSeek,
  title={DeepSeek-R1: Incentivizing Reasoning Capability in LLMs via Reinforcement Learning}, 
  author={DeepSeek-AI},
  year={2025},
  eprint={2501.12948},
  archivePrefix={arXiv},
  primaryClass={cs.CL},
  url={https://arxiv.org/abs/2501.12948}, 
}

@inproceedings{
      liu2024autodan,
      title={AutoDAN: Generating Stealthy Jailbreak Prompts on Aligned Large Language Models},
      author={Xiaogeng Liu and Nan Xu and Muhao Chen and Chaowei Xiao},
      booktitle={{ICLR}},
      year={2024},
}

@inproceedings{adv_training,
  author       = {Aleksander Madry and
                  Aleksandar Makelov and
                  Ludwig Schmidt and
                  Dimitris Tsipras and
                  Adrian Vladu},
  title        = {Towards Deep Learning Models Resistant to Adversarial Attacks},
  booktitle    = {6th International Conference on Learning Representations, {ICLR}},
  year         = {2018},
}

@inproceedings{WaNet,
  author    = {Tuan Anh Nguyen and
               Anh Tuan Tran},
  title     = {{WaNet - Imperceptible Warping-based Backdoor Attack}},
  booktitle = {ICLR},
  year      = {2021},
}

@inproceedings{PaudiceML18,
  author    = {Andrea Paudice and
               Luis Mu{\~{n}}oz{-}Gonz{\'{a}}lez and
               Emil C. Lupu},
  title     = {Label Sanitization Against Label Flipping Poisoning Attacks},
  booktitle = {Proc. {ECML} {PKDD} Workshops},
  year      = {2018},
}

@inproceedings{in-flight,
  author       = {Xi Li and
                  Zhen Xiang and
                  David J. Miller and
                  George Kesidis},
  title        = {Test-Time Detection of Backdoor Triggers for Poisoned Deep Neural  Networks},
  booktitle    = {IEEE ICASSP},
  year         = {2022},
}

@article{BNA,
title = {Correcting the distribution of batch normalization signals for Trojan mitigation},
journal = {Neurocomputing},
volume = {614},
pages = {128752},
year = {2025},
issn = {0925-2312},
doi = {https://doi.org/10.1016/j.neucom.2024.128752},
url = {https://www.sciencedirect.com/science/article/pii/S0925231224015236},
author = {Xi Li and Zhen Xiang and David J. Miller and George Kesidis},
}

@article{yin2024survey,
  title={A survey on multimodal large language models},
  author={Yin, Shukang and Fu, Chaoyou and Zhao, Sirui and Li, Ke and Sun, Xing and Xu, Tong and Chen, Enhong},
  journal={National Science Review},
  volume={11},
  number={12},
  year={2024}
}

@article{baltruvsaitis2018multimodal,
  title={Multimodal machine learning: A survey and taxonomy},
  author={Baltru{\v{s}}aitis, Tadas and Ahuja, Chaitanya and Morency, Louis-Philippe},
  journal={IEEE transactions on pattern analysis and machine intelligence},
  volume={41},
  number={2},
  pages={423--443},
  year={2018},
  publisher={IEEE}
}

@article{xu2023multimodal,
  title={Multimodal learning with transformers: A survey},
  author={Xu, Peng and Zhu, Xiatian and Clifton, David A},
  journal={IEEE Transactions on Pattern Analysis and Machine Intelligence},
  volume={45},
  number={10},
  pages={12113--12132},
  year={2023},
  publisher={IEEE}
}

@article{zhu2023multimodal,
  title={Multimodal sentiment analysis based on fusion methods: A survey},
  author={Zhu, Linan and Zhu, Zhechao and Zhang, Chenwei and Xu, Yifei and Kong, Xiangjie},
  journal={Information Fusion},
  volume={95},
  pages={306--325},
  year={2023},
  publisher={Elsevier}
}

@article{wang2023large,
  title={Large-scale multi-modal pre-trained models: A comprehensive survey},
  author={Wang, Xiao and Chen, Guangyao and Qian, Guangwu and Gao, Pengcheng and Wei, Xiao-Yong and Wang, Yaowei and Tian, Yonghong and Gao, Wen},
  journal={Machine Intelligence Research},
  volume={20},
  number={4},
  pages={447--482},
  year={2023},
  publisher={Springer}
}

@inproceedings{BadWord,
  author       = {Linyang Li and
                  Demin Song and
                  Xiaonan Li and
                  Jiehang Zeng and
                  Ruotian Ma and
                  Xipeng Qiu},
  title        = {Backdoor Attacks on Pre-trained Models by Layerwise Weight Poisoning},
  booktitle    = {{EMNLP}},
  year         = {2021},
}

@article{AddSent,
  author       = {Jiazhu Dai and
                  Chuanshuai Chen and
                  Yufeng Li},
  title        = {A Backdoor Attack Against LSTM-Based Text Classification Systems},
  journal      = {{IEEE} Access},
  year         = {2019},
}

@inproceedings{NC,
  author    = {Bolun Wang and
               Yuanshun Yao and
               Shawn Shan and
               Huiying Li and
               Bimal Viswanath and
               Haitao Zheng and
               Ben Y. Zhao},
  title     = {{Neural Cleanse: Identifying and Mitigating Backdoor Attacks in Neural Networks}},
  booktitle = {2019 {IEEE} Symposium on Security and Privacy},
  year      = {2019},
}

@inproceedings{STRIP,
  author    = {Yansong Gao and
               Change Xu and
               Derui Wang and
               Shiping Chen and
               Damith Chinthana Ranasinghe and
               Surya Nepal},
  title     = {{STRIP:} a defence against trojan attacks on deep neural networks},
  booktitle = {ACSAC},
  year      = {2019},
}

@inproceedings{FP,
author    = {Kang Liu and
           Brendan Dolan{-}Gavitt and
           Siddharth Garg},
title     = {{Fine-Pruning: Defending Against Backdooring Attacks on Deep Neural Networks}},
booktitle = {RAID},
year      = {2018},
}

@inproceedings{SS,
  author    = {Brandon Tran and
               Jerry Li and
               Aleksander Madry},
  title     = {{Spectral Signatures in Backdoor Attacks}},
  booktitle = {NeurIPS},
  year      = {2018},
}

@inproceedings{NAD,
  author    = {Yige Li and
               Xixiang Lyu and
               Nodens Koren and
               Lingjuan Lyu and
               Bo Li and
               Xingjun Ma},
  title     = {{Neural Attention Distillation: Erasing Backdoor Triggers from Deep
               Neural Networks}},
  booktitle = {ICLR},
  year      = {2021},
}

@inproceedings{hypergrad,
  title={{Adversarial Unlearning of Backdoors via Implicit Hypergradient}},
  author={Zeng, Yi and Chen, Si and Park, Won and Mao, Zhuoqing and Jin, Ming and Jia, Ruoxi},
  booktitle={ICLR},
  year={2022}
}

@inproceedings{FA,
  author       = {Wenxiao Wang and
                  Alexander Levine and
                  Soheil Feizi},
  title        = {Improved Certified Defenses against Data Poisoning with (Deterministic) Finite Aggregation},
  booktitle    = {{ICML}},
  year         = {2022},
}

@inproceedings{DPA,
  author       = {Alexander Levine and
                  Soheil Feizi},
  title        = {Deep Partition Aggregation: Provable Defenses against General Poisoning Attacks},
  booktitle    = {{ICLR}},
  year         = {2021},
}

@InProceedings{random_smoothing,
  title = 	 {Certified Adversarial Robustness via Randomized Smoothing},
  author =       {Cohen, Jeremy and Rosenfeld, Elan and Kolter, Zico},
  booktitle = 	 {Proceedings of the 36th International Conference on Machine Learning},
  pages = 	 {1310--1320},
  year = 	 {2019},
  editor = 	 {Chaudhuri, Kamalika and Salakhutdinov, Ruslan},
  volume = 	 {97},
  series = 	 {Proceedings of Machine Learning Research},
}

@misc{Llama3,
title={The Llama 3 Herd of Models}, 
author={Meta Llama3 team},
year={2024},
eprint={2407.21783},
archivePrefix={arXiv},
primaryClass={cs.AI},
url={https://arxiv.org/abs/2407.21783}, 
}

@inproceedings{ZhaoPDYLCL23,
  author       = {Yunqing Zhao and
                  Tianyu Pang and
                  Chao Du and
                  Xiao Yang and
                  Chongxuan Li and
                  Ngai{-}Man Cheung and
                  Min Lin},
  title        = {On Evaluating Adversarial Robustness of Large Vision-Language Models},
  booktitle    = {NeurIPS},
  year         = {2023},
}

@inproceedings{LuoGL024,
  author       = {Haochen Luo and
                  Jindong Gu and
                  Fengyuan Liu and
                  Philip Torr},
  title        = {An Image Is Worth 1000 Lies: Transferability of Adversarial Images
                  across Prompts on Vision-Language Models},
  booktitle    = {ICLR},
  year         = {2024},
}

@inproceedings{0003YZDZLCWM23,
  author       = {Ziyi Yin and
                  Muchao Ye and
                  Tianrong Zhang and
                  Tianyu Du and
                  Jinguo Zhu and
                  Han Liu and
                  Jinghui Chen and
                  Ting Wang and
                  Fenglong Ma},
  title        = {{VLATTACK:} Multimodal Adversarial Attacks on Vision-Language Tasks
                  via Pre-trained Models},
  booktitle    = {NeurIPS},
  year         = {2023},
}

@inproceedings{GaoBGX00024,
  author       = {Kuofeng Gao and
                  Yang Bai and
                  Jindong Gu and
                  Shu{-}Tao Xia and
                  Philip Torr and
                  Zhifeng Li and
                  Wei Liu},
  title        = {Inducing High Energy-Latency of Large Vision-Language Models with
                  Verbose Images},
  booktitle    = {ICLR},
  year         = {2024},
}

@inproceedings{WangDZQLFWL24,
  author       = {Haodi Wang and
                  Kai Dong and
                  Zhilei Zhu and
                  Haotong Qin and
                  Aishan Liu and
                  Xiaolin Fang and
                  Jiakai Wang and
                  Xianglong Liu},
  title        = {Transferable Multimodal Attack on Vision-Language Pre-training Models},
  booktitle    = {{IEEE} {S \& P}},
  year         = {2024},
}

@inproceedings{WangLQCJX24,
  author       = {Yubo Wang and
                  Chaohu Liu and
                  Yanqiu Qu and
                  Haoyu Cao and
                  Deqiang Jiang and
                  Linli Xu},
  title        = {Break the Visual Perception: Adversarial Attacks Targeting Encoded
                  Visual Tokens of Large Vision-Language Models},
  booktitle    = {{ACM} {MM}},
  year         = {2024},
}

@inproceedings{BaileyO0E24,
  author       = {Luke Bailey and
                  Euan Ong and
                  Stuart Russell and
                  Scott Emmons},
  title        = {Image Hijacks: Adversarial Images can Control Generative Models at
                  Runtime},
  booktitle    = {{ICML}},
  year         = {2024},
}

@inproceedings{Schlarmann023,
  author       = {Christian Schlarmann and
                  Matthias Hein},
  title        = {On the Adversarial Robustness of Multi-Modal Foundation Models},
  booktitle    = {{ICCV} - Workshops},
  year         = {2023},
}

@inproceedings{LiuYQ0FT0024,
  author       = {Daizong Liu and
                  Mingyu Yang and
                  Xiaoye Qu and
                  Pan Zhou and
                  Xiang Fang and
                  Keke Tang and
                  Yao Wan and
                  Lichao Sun},
  title        = {Pandora's Box: Towards Building Universal Attackers against Real-World
                  Large Vision-Language Models},
  booktitle    = {NeurIPS},
  year         = {2024},
}

@inproceedings{WalmerSSSJ22,
  author       = {Matthew Walmer and
                  Karan Sikka and
                  Indranil Sur and
                  Abhinav Shrivastava and
                  Susmit Jha},
  title        = {Dual-Key Multimodal Backdoors for Visual Question Answering},
  booktitle    = {CVPR},
  year         = {2022},
}

@inproceedings{BaiGMX0024,
  author       = {Jiawang Bai and
                  Kuofeng Gao and
                  Shaobo Min and
                  Shu{-}Tao Xia and
                  Zhifeng Li and
                  Wei Liu},
  title        = {BadCLIP: Trigger-Aware Prompt Learning for Backdoor Attacks on {CLIP}},
  booktitle    = {CVPR},
  year         = {2024},
}

@article{abs-2402-08577,
  author       = {Dong Lu and
                  Tianyu Pang and
                  Chao Du and
                  Qian Liu and
                  Xianjun Yang and
                  Min Lin},
  title        = {Test-Time Backdoor Attacks on Multimodal Large Language Models},
  journal      = {arXiv: 2402.08577},
  year         = {2024},
}

@inproceedings{LiangLPDLZCT25,
  author       = {Siyuan Liang and
                  Jiawei Liang and
                  Tianyu Pang and
                  Chao Du and
                  Aishan Liu and
                  Mingli Zhu and
                  Xiaochun Cao and
                  Dacheng Tao},
  title        = {Revisiting Backdoor Attacks against Large Vision-Language Models from
                  Domain Shift},
  booktitle    = {{CVPR}},
  pages        = {9477--9486},
  publisher    = {Computer Vision Foundation / {IEEE}},
  year         = {2025},
}

@inproceedings{jailbreak_adv,
  author       = {Erfan Shayegani and
                  Yue Dong and
                  Nael B. Abu{-}Ghazaleh},
  title        = {Jailbreak in pieces: Compositional Adversarial Attacks on Multi-Modal
                  Language Models},
  booktitle    = {{ICLR}},
  year         = {2024},
}

@inproceedings{ImgTrojan,
  author       = {Xijia Tao and
                  Shuai Zhong and
                  Lei Li and
                  Qi Liu and
                  Lingpeng Kong},
  title        = {ImgTrojan: Jailbreaking Vision-Language Models with {ONE} Image},
  booktitle    = {{NAACL}},
  year         = {2025},
}

@inproceedings{FigStep,
  author       = {Yichen Gong and
                  Delong Ran and
                  Jinyuan Liu and
                  Conglei Wang and
                  Tianshuo Cong and
                  Anyu Wang and
                  Sisi Duan and
                  Xiaoyun Wang},
  title        = {FigStep: Jailbreaking Large Vision-Language Models via Typographic
                  Visual Prompts},
  booktitle    = {{AAAI}},
  year         = {2025},
}

@inproceedings{QiHP0WM24,
  author       = {Xiangyu Qi and
                  Kaixuan Huang and
                  Ashwinee Panda and
                  Peter Henderson and
                  Mengdi Wang and
                  Prateek Mittal},
  title        = {Visual Adversarial Examples Jailbreak Aligned Large Language Models},
  booktitle    = {{AAAI}},
  year         = {2024},
}

@article{BadNet,
  author    = {Tianyu Gu and
               Kang Liu and
               Brendan Dolan{-}Gavitt and
               Siddharth Garg},
  title     = {{BadNets: Evaluating Backdooring Attacks on Deep Neural Networks}},
  journal   = {{IEEE} Access},
  year      = {2019},
}

@article{Targeted-Backdoor,
  author    = {Xinyun Chen and
               Chang Liu and
               Bo Li and
               Kimberly Lu and
               Dawn Song},
  title     = {{Targeted Backdoor Attacks on Deep Learning Systems Using Data Poisoning}},
  journal   = {arXiv:1712.05526},
  year      = {2017},
}

@inproceedings{HiddenTrigger,
  author    = {Aniruddha Saha and
               Akshayvarun Subramanya and
               Hamed Pirsiavash},
  title     = {{Hidden Trigger Backdoor Attacks}},
  booktitle = {AAAI},
  year      = {2020},
}

@inproceedings{input-aware,
  author       = {Tuan Anh Nguyen and
                  Anh Tuan Tran},
  title        = {Input-Aware Dynamic Backdoor Attack},
  booktitle    = {NeurIPS},
  year         = {2020},
}

@inproceedings{QiCZLLS21,
  author       = {Fanchao Qi and
                  Yangyi Chen and
                  Xurui Zhang and
                  Mukai Li and
                  Zhiyuan Liu and
                  Maosong Sun},
  title        = {Mind the Style of Text! Adversarial and Backdoor Attacks Based on
                  Text Style Transfer},
  booktitle    = {EMNLP},
  year         = {2021}
}

@inproceedings{XiaoXE12,
  author       = {Han Xiao and
                  Huang Xiao and
                  Claudia Eckert},
  title        = {Adversarial Label Flips Attack on Support Vector Machines},
  booktitle    = {{ECAI}},
  volume       = {242},
  pages        = {870--875},
  year         = {2012},
}

@article{ZhangCZL21,
  author       = {Hongpo Zhang and
                  Ning Cheng and
                  Yang Zhang and
                  Zhanbo Li},
  title        = {Label flipping attacks against Naive Bayes on spam filtering systems},
  journal      = {Appl. Intell.},
  volume       = {51},
  number       = {7},
  pages        = {4503--4514},
  year         = {2021},
}

@inproceedings{li2021anti,
  title={{Anti-Backdoor Learning: Training Clean Models on Poisoned Data}},
  author={Li, Yige and Lyu, Xixiang and Koren, Nodens and Lyu, Lingjuan and Li, Bo and Ma, Xingjun},
  booktitle={NeurIPS},
  year={2021}
}

@inproceedings{DiakonikolasKK019,
  author    = {Ilias Diakonikolas and
               Gautam Kamath and
               Daniel Kane and
               Jerry Li and
               Jacob Steinhardt and
               Alistair Stewart},
  title     = {Sever: {A} Robust Meta-Algorithm for Stochastic Optimization},
  booktitle = {ICML},
  year      = {2019},
}

@InProceedings{trim_loss,
	title = 	 {{Learning with Bad Training Data via Iterative Trimmed Loss Minimization}},
	author =       {Shen, Yanyao and Sanghavi, Sujay},
	booktitle = 	 {ICML},
	pages = 	 {5739--5748},
	year = 	 {2019},
}

@inproceedings{UNICORN,
  author       = {Zhenting Wang and
                  Kai Mei and
                  Juan Zhai and
                  Shiqing Ma},
  title        = {{UNICORN:} {A} Unified Backdoor Trigger Inversion Framework},
  booktitle    = {{ICLR}},
  year         = {2023},
}

@inproceedings{FGSM,
  author       = {Ian J. Goodfellow and
                  Jonathon Shlens and
                  Christian Szegedy},
  title        = {Explaining and Harnessing Adversarial Examples},
  booktitle    = {{ICLR}},
  year         = {2015},
}

@inproceedings{PGD,
  author       = {Aleksander Madry and
                  Aleksandar Makelov and
                  Ludwig Schmidt and
                  Dimitris Tsipras and
                  Adrian Vladu},
  title        = {Towards Deep Learning Models Resistant to Adversarial Attacks},
  booktitle    = {{ICLR}},
  year         = {2018},
}

@inproceedings{CW,
  author       = {Nicholas Carlini and
                  David A. Wagner},
  title        = {Towards Evaluating the Robustness of Neural Networks},
  booktitle    = {{IEEE} Symposium on Security and Privacy},
  pages        = {39--57},
  year         = {2017},
}

@inproceedings{ZOO,
  author       = {Pin{-}Yu Chen and
                  Huan Zhang and
                  Yash Sharma and
                  Jinfeng Yi and
                  Cho{-}Jui Hsieh},
  title        = {{ZOO:} Zeroth Order Optimization Based Black-box Attacks to Deep Neural
                  Networks without Training Substitute Models},
  booktitle    = {AISec@CCS},
  year         = {2017},
}

@inproceedings{LiCWC19,
  author       = {Bai Li and
                  Changyou Chen and
                  Wenlin Wang and
                  Lawrence Carin},
  title        = {Certified Adversarial Robustness with Additive Noise},
  booktitle    = {NeurIPS},
  pages        = {9459--9469},
  year         = {2019},
}

@inproceedings{JonesDRS23,
  author       = {Erik Jones and
                  Anca D. Dragan and
                  Aditi Raghunathan and
                  Jacob Steinhardt},
  title        = {Automatically Auditing Large Language Models via Discrete Optimization},
  booktitle    = {{ICML}},
  volume       = {202},
  pages        = {15307--15329},
  year         = {2023},
}

@article{abs-2307-15043,
  author       = {Andy Zou and
                  Zifan Wang and
                  J. Zico Kolter and
                  Matt Fredrikson},
  title        = {Universal and Transferable Adversarial Attacks on Aligned Language
                  Models},
  journal      = {CoRR},
  volume       = {abs/2307.15043},
  year         = {2023},
}

@article{abs-2310-15140,
  author       = {Sicheng Zhu and Ruiyi Zhang and Bang An and Gang Wu and Joe Barrow and Zichao Wang and Furong Huang and Ani Nenkova and Tong Sun},
  title        = {AutoDAN: Interpretable Gradient-Based Adversarial Attacks on Large Language Models},
  journal      = {CoRR},
  volume       = {abs/2310.15140},
  year         = {2023},
}

@inproceedings{GuoYZQ024,
  author       = {Xingang Guo and
                  Fangxu Yu and
                  Huan Zhang and
                  Lianhui Qin and
                  Bin Hu},
  title        = {COLD-Attack: Jailbreaking LLMs with Stealthiness and Controllability},
  booktitle    = {{ICML}},
  year         = {2024},
}

@article{abs-2401-17256,
  author       = {Xuandong Zhao and
                  Xianjun Yang and
                  Tianyu Pang and
                  Chao Du and
                  Lei Li and
                  Yu{-}Xiang Wang and
                  William Yang Wang},
  title        = {Weak-to-Strong Jailbreaking on Large Language Models},
  journal      = {CoRR},
  volume       = {abs/2401.17256},
  year         = {2024},
}

@inproceedings{Qi0XC0M024,
  author       = {Xiangyu Qi and
                  Yi Zeng and
                  Tinghao Xie and
                  Pin{-}Yu Chen and
                  Ruoxi Jia and
                  Prateek Mittal and
                  Peter Henderson},
  title        = {Fine-tuning Aligned Language Models Compromises Safety, Even When
                  Users Do Not Intend To!},
  booktitle    = {{ICLR}},
  year         = {2024},
}

@inproceedings{ZhanFBGHK24,
  author       = {Qiusi Zhan and
                  Richard Fang and
                  Rohan Bindu and
                  Akul Gupta and
                  Tatsunori Hashimoto and
                  Daniel Kang},
  title        = {Removing {RLHF} Protections in {GPT-4} via Fine-Tuning},
  booktitle    = {{NAACL}},
  pages        = {681--687},
  year         = {2024},
}

@article{abs-2310-02949,
  author       = {Xianjun Yang and
                  Xiao Wang and
                  Qi Zhang and
                  Linda R. Petzold and
                  William Yang Wang and
                  Xun Zhao and
                  Dahua Lin},
  title        = {Shadow Alignment: The Ease of Subverting Safely-Aligned Language Models},
  journal      = {CoRR},
  volume       = {abs/2310.02949},
  year         = {2023},
}

@article{abs-2305-14950,
  author       = {Jiongxiao Wang and
                  Zichen Liu and
                  Keun Hee Park and
                  Muhao Chen and
                  Chaowei Xiao},
  title        = {Adversarial Demonstration Attacks on Large Language Models},
  journal      = {CoRR},
  volume       = {abs/2305.14950},
  year         = {2023},
}

@article{abs-2310-06387,
  author       = {Zeming Wei and
                  Yifei Wang and
                  Yisen Wang},
  title        = {Jailbreak and Guard Aligned Language Models with Only Few In-Context
                  Demonstrations},
  journal      = {CoRR},
  volume       = {abs/2310.06387},
  year         = {2023},
}

@article{abs-2311-03191,
  author       = {Xuan Li and
                  Zhanke Zhou and
                  Jianing Zhu and
                  Jiangchao Yao and
                  Tongliang Liu and
                  Bo Han},
  title        = {DeepInception: Hypnotize Large Language Model to Be Jailbreaker},
  journal      = {CoRR},
  volume       = {abs/2311.03191},
  year         = {2023},
}

@inproceedings{YuanJW0H0T24,
  author       = {Youliang Yuan and
                  Wenxiang Jiao and
                  Wenxuan Wang and
                  Jen{-}tse Huang and
                  Pinjia He and
                  Shuming Shi and
                  Zhaopeng Tu},
  title        = {{GPT-4} Is Too Smart To Be Safe: Stealthy Chat with LLMs via Cipher},
  booktitle    = {{ICLR}},
  year         = {2024},
}

@inproceedings{Deng_2024, 
   title={MASTERKEY: Automated Jailbreaking of Large Language Model Chatbots},
   booktitle={{NDSS}},
   author={Deng, Gelei and Liu, Yi and Li, Yuekang and Wang, Kailong and Zhang, Ying and Li, Zefeng and Wang, Haoyu and Zhang, Tianwei and Liu, Yang},
   year={2024},
}

@article{abs-2308-14132,
  author       = {Gabriel Alon and
                  Michael Kamfonas},
  title        = {Detecting Language Model Attacks with Perplexity},
  journal      = {CoRR},
  volume       = {abs/2308.14132},
  year         = {2023},
}

@article{abs-2309-00614,
  author       = {Neel Jain and
                  Avi Schwarzschild and
                  Yuxin Wen and
                  Gowthami Somepalli and
                  John Kirchenbauer and
                  Ping{-}yeh Chiang and
                  Micah Goldblum and
                  Aniruddha Saha and
                  Jonas Geiping and
                  Tom Goldstein},
  title        = {Baseline Defenses for Adversarial Attacks Against Aligned Language
                  Models},
  journal      = {CoRR},
  volume       = {abs/2309.00614},
  year         = {2023},
}

@inproceedings{Ouyang0JAWMZASR22,
  author       = {Long Ouyang and
                  Jeffrey Wu and
                  Xu Jiang and
                  Diogo Almeida and
                  Carroll L. Wainwright and
                  Pamela Mishkin and
                  Chong Zhang and
                  Sandhini Agarwal and
                  Katarina Slama and
                  Alex Ray and
                  John Schulman and
                  Jacob Hilton and
                  Fraser Kelton and
                  Luke Miller and
                  Maddie Simens and
                  Amanda Askell and
                  Peter Welinder and
                  Paul F. Christiano and
                  Jan Leike and
                  Ryan Lowe},
  title        = {Training language models to follow instructions with human feedback},
  booktitle    = {{NeurIPS}},
  year         = {2022},
}

@inproceedings{SunSZZCCYG23,
  author       = {Zhiqing Sun and
                  Yikang Shen and
                  Qinhong Zhou and
                  Hongxin Zhang and
                  Zhenfang Chen and
                  David D. Cox and
                  Yiming Yang and
                  Chuang Gan},
  title        = {Principle-Driven Self-Alignment of Language Models from Scratch with
                  Minimal Human Supervision},
  booktitle    = {{NeurIPS}},
  year         = {2023},
}

@inproceedings{0001SARJH024,
  author       = {Federico Bianchi and
                  Mirac Suzgun and
                  Giuseppe Attanasio and
                  Paul R{\"{o}}ttger and
                  Dan Jurafsky and
                  Tatsunori Hashimoto and
                  James Zou},
  title        = {Safety-Tuned LLaMAs: Lessons From Improving the Safety of Large Language
                  Models that Follow Instructions},
  booktitle    = {{ICLR}},
  publisher    = {OpenReview.net},
  year         = {2024},
}

@inproceedings{DengWFDW023,
  author       = {Boyi Deng and
                  Wenjie Wang and
                  Fuli Feng and
                  Yang Deng and
                  Qifan Wang and
                  Xiangnan He},
  title        = {Attack Prompt Generation for Red Teaming and Defending Large Language
                  Models},
  booktitle    = {Findings of {EMNLP}},
  year         = {2023},
}

@inproceedings{YuLLCXZ24,
  author       = {Zhiyuan Yu and
                  Xiaogeng Liu and
                  Shunning Liang and
                  Zach Cameron and
                  Chaowei Xiao and
                  Ning Zhang},
  title        = {Don't Listen To Me: Understanding and Exploring Jailbreak Prompts
                  of Large Language Models},
  booktitle    = {{USENIX} Security},
  year         = {2024},
}

@inproceedings{XuYSSW0GH24,
  author       = {Yuancheng Xu and
                  Jiarui Yao and
                  Manli Shu and
                  Yanchao Sun and
                  Zichu Wu and
                  Ning Yu and
                  Tom Goldstein and
                  Furong Huang},
  title        = {Shadowcast: Stealthy Data Poisoning Attacks Against Vision-Language
                  Models},
  booktitle    = {NeurIPS},
  year         = {2024},
}

@inproceedings{bagdasaryan2024adversarial,
  title={Adversarial illusions in Multi-Modal embeddings},
  author={Bagdasaryan, Eugene and Jha, Rishi and Shmatikov, Vitaly and Zhang, Tingwei},
  booktitle={33rd USENIX Security Symposium (USENIX Security 24)},
  pages={3009--3025},
  year={2024}
}

@inproceedings{tu2024many,
  title={How many are in this image a safety evaluation benchmark for vision llms},
  author={Tu, Haoqin and Cui, Chenhang and Wang, Zijun and Zhou, Yiyang and Zhao, Bingchen and Han, Junlin and Zhou, Wangchunshu and Yao, Huaxiu and Xie, Cihang},
  booktitle={European Conference on Computer Vision},
  pages={37--55},
  year={2024},
  organization={Springer}
}

@inproceedings{huangx,
  title={X-Transfer Attacks: Towards Super Transferable Adversarial Attacks on CLIP},
  author={Huang, Hanxun and Erfani, Sarah Monazam and Li, Yige and Ma, Xingjun and Bailey, James},
  booktitle={Forty-second International Conference on Machine Learning},
  year={2025}
}

@inproceedings{zhang2025anyattack,
  title={AnyAttack: Towards Large-scale Self-supervised Adversarial Attacks on Vision-language Models},
  author={Zhang, Jiaming and Ye, Junhong and Ma, Xingjun and Li, Yige and Yang, Yunfan and Chen, Yunhao and Sang, Jitao and Yeung, Dit-Yan},
  booktitle={{CVPR}},
  pages={19900--19909},
  year={2025}
}

@article{han2023ot,
  title={Ot-attack: Enhancing adversarial transferability of vision-language models via optimal transport optimization},
  author={Han, Dongchen and Jia, Xiaojun and Bai, Yang and Gu, Jindong and Liu, Yang and Cao, Xiaochun},
  journal={arXiv preprint arXiv:2312.04403},
  year={2023}
}

@inproceedings{lu2023set,
  title={Set-level guidance attack: Boosting adversarial transferability of vision-language pre-training models},
  author={Lu, Dong and Wang, Zhiqiang and Wang, Teng and Guan, Weili and Gao, Hongchang and Zheng, Feng},
  booktitle={Proceedings of the IEEE/CVF International Conference on Computer Vision},
  pages={102--111},
  year={2023}
}

@article{zhang2024b,
  title={B-avibench: Towards evaluating the robustness of large vision-language model on black-box adversarial visual-instructions},
  author={Zhang, Hao and Shao, Wenqi and Liu, Hong and Ma, Yongqiang and Luo, Ping and Qiao, Yu and Zheng, Nanning and Zhang, Kaipeng},
  journal={IEEE Transactions on Information Forensics and Security},
  year={2024},
  publisher={IEEE}
}

@inproceedings{cai2025imperceptible,
  title={Imperceptible Transfer Attack on Large Vision-Language Models},
  author={Cai, Xiaowen and Liu, Daizong and Guan, Runwei and Zhou, Pan},
  booktitle={ICASSP 2025-2025 IEEE International Conference on Acoustics, Speech and Signal Processing (ICASSP)},
  pages={1--5},
  year={2025},
  organization={IEEE}
}
\bibliographystyle{iclr2026_conference}

\end{document}